\documentclass{article}

\usepackage{mathtools}
\usepackage{amsthm}

\usepackage{microtype}
\usepackage{fourier}
\usepackage{bm}

\usepackage[letterpaper]{geometry}
\usepackage[affil-it]{authblk}

\usepackage[backend = biber, style = alphabetic, maxbibnames = 6]{biblatex}

\usepackage{enumitem}

\setenumerate{label = (\alph*), ref=\thelemma(\alph*)}

\usepackage[dvipsnames]{xcolor}
\usepackage{graphicx}

\usepackage{hyperref}

\newcommand*{\Z}{\mathbb{Z}}

\newcommand*{\R}{\mathbb{R}}

\newcommand*{\E}{\mathbb{E}} % for expectations

\newcommand*{\bbE}{\mathbb{E}}

\newcommand*{\bbP}{\mathbb{P}}

\newcommand*{\bbX}{\mathbb{X}}

\newcommand*{\bfU}{\mathbf{U}}
\newcommand*{\bfV}{\mathbf{V}}

\newcommand*{\bfX}{\mathbf{X}}
\newcommand*{\bfY}{\mathbf{Y}}

\newcommand*{\bfp}{\mathbf{p}}

\newcommand*{\bfu}{\mathbf{u}}
\newcommand*{\bfv}{\mathbf{v}}
\newcommand*{\bfw}{\mathbf{w}}
\newcommand*{\bfx}{\mathbf{x}}
\newcommand*{\bfy}{\mathbf{y}}
\newcommand*{\bfz}{\mathbf{z}}

\newcommand*{\bfeps}{\bm{\eps}}

\newcommand*{\bfzero}{\mathbf{0}}

\newcommand*{\calW}{\mathcal{W}}

\newcommand*{\inv}{^{-1}}
\newcommand*{\pinv}{^{+}}

\newcommand*{\tran}{^{\mathsf{t}}}

\newcommand*{\eps}{\varepsilon}

\newcommand*{\sigmax}{\sigma_{\mathrm{max}}}
\newcommand*{\sigmin}{\sigma_{\mathrm{min}}}
\newcommand*{\sigminp}{\sigma_{\mathrm{min}}^+}

\newcommand*{\bigmid}{\ \big\vert\ }
\newcommand*{\Bigmid}{\ \Big\vert\ }

\DeclarePairedDelimiter{\abs}{\lvert}{\rvert}

\DeclarePairedDelimiter{\norm}{\lVert}{\rVert}
\DeclarePairedDelimiter{\ceil}{\lceil}{\rceil}
\DeclarePairedDelimiter{\floor}{\lfloor}{\rfloor}

\newcommand*{\Ber}{\mathrm{Ber}}
\newcommand*{\Cov}{\mathrm{Cov}}

\newcommand*{\Tr}{\mathrm{Tr}}

\newcommand*{\id}{\mathrm{id}}

\newcommand*{\rank}{\mathrm{rank}}

\newcommand*{\rmd}{\mathrm{d}}
\newcommand*{\op}{\mathrm{op}}

\newcommand{\weight}{\bfw}

\newcommand{\whweight}{\widehat{\bfw}}
\newcommand{\whX}{\widehat{X}}
\newcommand{\whbbX}{\widehat{\bbX}}
\newcommand{\whY}{\widehat{\bfY}}

\newcommand{\whu}{\widehat{\bfu}}
\newcommand{\whv}{\widehat{\bfv}}

\newcommand{\starweight}{\bfw_{*}}

\newcommand{\Slin}{S^{\mathrm{lin}}}
\newcommand{\Sint}{S^{\mathrm{int}}}

\newtheorem{lemma}{Lemma}[section]

\newtheorem{theorem}[lemma]{Theorem}

\newtheorem{assumption}[lemma]{Assumption}

\title{The Interplay of Statistics and Noisy Optimization: Learning Linear
Predictors with Random Data Weights}

\author[1]{Gabriel Clara
  \footnote{Work chiefly conducted while based at the University of Twente.}
  \footnote{The authors thank Johannes Schmidt-Hieber and Sophie Langer for
    comments and productive conversations.

    \begin{minipage}{0.05\textwidth}
      \includegraphics[height = 2.5\baselineskip]{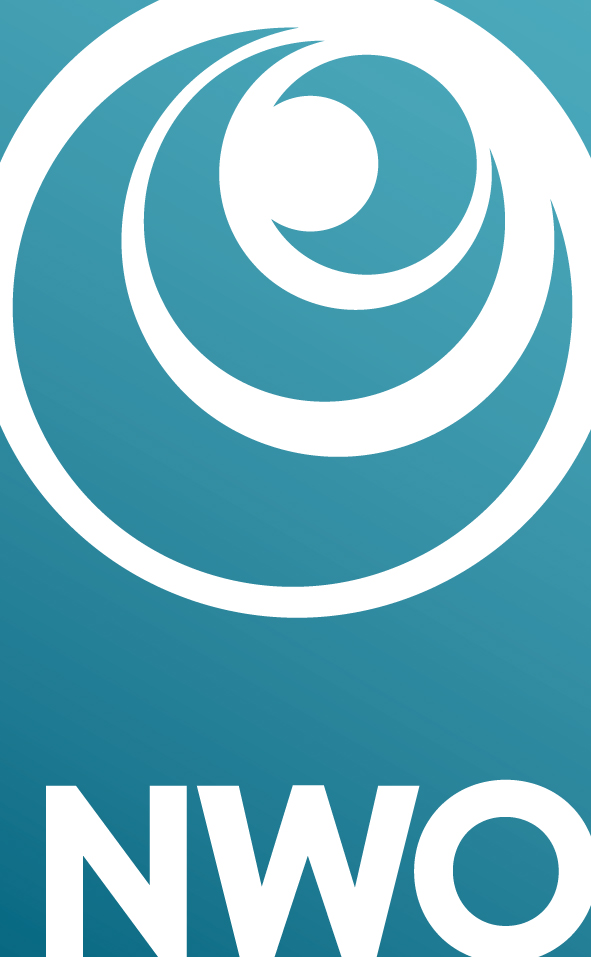}
    \end{minipage} \hfill \begin{minipage}{0.93\textwidth}
      This publication is part of the project \textit{Statistical foundation for
      multilayer neural networks} (project number VI.Vidi.192.021 of the Vidi
      ENW programme) financed by the Dutch Research Council (NWO).
    \end{minipage}}
}
\author[2]{Yazan Mash'al
  \protect\footnotemark[1] \protect\footnotemark[2] }

\affil[1]{Simons Institute for the Theory of Computing, University of
California, Berkeley}
\affil[2]{Institute of Applied Mathematics, Delft University of Technology}

\begin{document}

\maketitle

\begin{abstract}
  We analyze gradient descent with randomly weighted data points in a linear
  regression model, under a generic weighting distribution. This includes
  various forms of stochastic gradient descent, importance sampling, but also
  extends to weighting distributions with arbitrary continuous values, thereby
  providing a unified framework to analyze the impact of various kinds of noise
  on the training trajectory. We characterize the implicit regularization
  induced through the random weighting, connect it with weighted linear
  regression, and derive non-asymptotic bounds for convergence in first and
  second moments. Leveraging geometric moment contraction, we also investigate
  the stationary distribution induced by the added noise. Based on these
  results, we discuss how specific choices of weighting distribution influence
  both the underlying optimization problem and statistical properties of the
  resulting estimator, as well as some examples for which weightings that lead
  to fast convergence cause bad statistical performance.
\end{abstract}

\section{Introduction}

Gradient descent with random weightings of the data points is a ubiquitous
method in the training of large machine learning models. For example, during
each iteration of training, stochastic gradient descent (SGD) randomly selects a
mini-batch of the data on which to compute the gradient, which may be
interpreted as a random weighting with binary outcomes. This yields
computational gains if the batch size is relatively small in comparison with the
entire data set. The most basic instance of SGD features a fixed batch size and
uniform inclusion probability for each data point, but some variants use random
batch sizes \cite{bieringer_kasieczka_et_al_2023} and/or weighted probabilities
\cite{needell_srebro_et_al_2014, needell_ward_2017, csiba_richtarik_2018}.

While these methods were invented to make otherwise computationally intractable
problems accessible, they also induce a regularizing effect on the problem
\cite{smith_dherin_et_al_2021, wu_su_2023}. This effect, termed implicit
regularization, has recently experienced interest in the context of
generalization \cite{zhang_bengio_et_al_2017, zhang_bengio_et_al_2021}. In
empirical risk minimization (ERM), generalization refers to the ability of an
estimate to accurately predict the labels of previously unseen data. Non-convex
risks, such as those associated with heavily over-parametrized deep neural
networks, may admit many qualitatively different local and global minima, only
some of which generalize. Implicit regularization through the chosen
optimization method has been put forward as one plausible explanation for the
effectiveness of ERM in producing large machine learning models that generalize
\cite{bartlett_montanari_et_al_2021}.

One of the ways that random weighting of the data points influences the computed
parameter estimates concerns the diffusion of the iterates through the parameter
space. As an example, consider the empirical risk \begin{align*}
  f(\bfw) = \dfrac{1}{n} \cdot \sum_{i = 1}^n f_i(\bfw),
\end{align*} with $f_i$ representing the loss on the $i$\textsuperscript{th}
data point. For a given initialization $\bfw_0$, the classical SGD recursion
takes the form $\bfw_{k + 1} = \bfw_k - \alpha \cdot \nabla f_{i_k}(\bfw_{k})$,
with $i_k \sim \mathrm{Unif}(1, \ldots, n)$ and step-size $\alpha > 0$. The
conditional expectation of $\nabla f_{i_k}$ matches $\nabla f$, so the long-run
distribution of $\bfw_k$ clusters around the critical points of $f$ under mild
assumptions \cite{azizian_iutzeler_et_al_2024}. The variance induced through the
random sampling of the $i_k$ may help the SGD algorithm in escaping sharp local
minima \cite{ibayashi_imaizumi_2023}, which leads to better generalization
guarantees \cite{haddouche_viallard_et_al_2025}.

In the example above, the added noise does not change the conditional
expectation of the gradient, but this may not always be the case. For biased
sampling of the $i_k$, the long-run distribution of the iterates will cluster
around the critical points of the weighted empirical loss, with each $f_i$
weighted according to the probability of $i_k = i$. Biased sampling may be
desirable if some of the observed data points hold a stronger influence on the
overall loss than others \cite{needell_srebro_et_al_2014, needell_ward_2017}. In
this case, the noise changes the underlying loss surface and hence the resulting
trajectory of the SGD iterates.

In the present article, we focus on linear regression as a well-understood model
and analyze the impact of random weightings on the underlying gradient descent
dynamics. While the linear model does not capture the complexities associated
with non-convex ERM, it serves as a useful starting point to characterize the
induced regularization. Further, SGD with biased sampling in this model features
a striking relationship with randomized algorithms for the solution of linear
systems \cite{needell_srebro_et_al_2014}. On the technical side, we adapt recent
progress in the analysis of linear regression with dropout
\cite{clara_langer_et_al_2024, li_schmidt-hieber_et_al_2024}, with the aim of
characterizing the interactions between gradient descent dynamics and the added
variance, as well as the resulting stationary distribution. Throughout, we
provide a unified analysis for generic random weightings, meaning the
distribution is not limited to discrete outcomes. Continuous weightings may be
desirable to down-weight outliers and increase robustness, both in classical
statistics \cite{holland_welsch_1977} and modern machine learning
\cite{de_brabanter_de_brabanter_2021, ren_zeng_et_al_2018}. Further,
continuously-valued random data weights relate to curriculum learning
\cite{bengio_weston_2009} and randomized sketching methods \cite{vempala_2004}.

The article is organized as follows: Section \ref{sec::grad_desc} gives a short
overview of gradient descent in the linear model and discusses the relationship
between weighted linear least squares (W-LLS) and random weightings. Section
\ref{sec::conv_res_iso} details our convergence analysis of randomly weighted
gradient descent. Section \ref{sec::spec_sim} discusses specific weighting
strategies and their influence on both optimization and statistical properties
of the algorithm. Section \ref{sec::disc} concludes with further discussion of
future research directions. The proofs of all numbered statements are deferred
to the appendices.

\subsection{Related Work}

The idea of approximating an iterative optimization algorithm via random
sampling finds its genesis in \cite{robbins_monro_1951}, often considered as
having founded the study of stochastic approximation methods. Motivated by
questions in machine learning, far ranging generalizations of classical
stochastic approximation results have recently been obtained
\cite{davis_drusvyatskiy_et_al_2020, fehrman_gess_et_al_2020,
mertikopoulos_hallak_et_al_2020, dereich_kassing_2024}. While we do not use
these abstract convergence theorems, our results may be seen as concrete
counterparts to such theorems that exploit the precise structure of the linear
model as much as possible.

As a widely used method, SGD has inspired many studies, so we only summarize the
most relevant directions of research. The batch size, as a tuning parameter,
controls the variance of the gradient estimators and hence directly influences
the training dynamics \cite{wilson_martinez_2003}. For a weight-tied non-linear
auto-encoder, \cite{ghosh_frei_et_al_2025} show that the limit reached by SGD
depends directly on the batch size. Biased sampling represents a further way to
alter the dynamics, which may lead to faster convergence and connects to the
randomized Kaczmarz method for linear systems \cite{needell_srebro_et_al_2014}.
One may also sample mini-batches with biased inclusion probabilities, with a
natural sampling rule being proposed in \cite{needell_ward_2017}. In a similar
setting, \cite{csiba_richtarik_2018} analyze the convergence of a stochastic
dual coordinate ascent algorithm with importance sampling.

Toy model analysis of regularization induced by optimization algorithms is an
active area of research. As shown in \cite{bartlett_long_et_al_2023},
sharpness-aware minimization forces the GD algorithm to asymptotically bounce
between two opposing sides of a valley when applied to a quadratic objective.
For diagonal linear networks trained with a stochastic version of SAM,
\cite{clara_langer_et_al_2025} compute the exact induced regularizer, which
leads to exponentially fast balancing of the weight matrices along the GD
trajectory. \cite{wu_bartlett_et_al_2025} provide excess risk bounds for early
stopping in logistic regression and investigate connections to explicit norm
regularization. \cite{clara_langer_et_al_2024} analyze the implicit and explicit
regularizing effect of dropout in linear regression and
\cite{li_schmidt-hieber_et_al_2024} prove uniqueness of the induced stationary
distribution and a quenched central limit theorem for the averaged iterates in
the same setting.

For constant step-sizes, noisy GD algorithms produce iterates that diffuse
through the parameter space indefinitely, unless the noise vanishes as the
iterates approach a stable point. The resulting long-run distribution of the
iterates reflects the geometry of the loss surface by clustering near critical
points and may be shown to visit flat regions with higher frequency
\cite{azizian_iutzeler_et_al_2024}. \cite{shalova_schlichting_et_al_2024}
exhibit the asymptotic constraints induced by arbitrary noise in the small
step-size limit. Ergodicity and asymptotic normality of the SGD iterate sequence
for a non-convex loss is shown in \cite{yu_balasubramanian_et_al_2021}.

\subsection{Notation}
\label{sec::not}

We let $A\tran$, $\Tr(A)$, and $A\pinv$ signify the transpose, trace, and
pseudo-inverse of a matrix $A$. The maximal and minimal singular values of a
matrix are denoted by $\sigmax(A)$ and $\sigmin(A)$, with $\sigminp(A)$
signifying the smallest non-zero singular value. For matrices of the same
dimension, $A \odot B$ gives the element-wise product $(A \odot B)_{ij} = A_{ij}
B_{ij}$.

Euclidean vectors are always written in boldface and endowed with the standard
norm $\norm{\bfv}_2^2 = \bfv\tran \bfv$. Given a matrix $A$ of suitable
dimension, $\bfv \in \ker(A)$ whenever $A \bfv = \bfzero$. The orthogonal
complement of $\ker(A)$ contains all vectors satisfying $\bfw\tran \bfv = 0$ for
every $\bfv \in \ker(A)$; we then also write $\bfv \perp \ker(A)$.

Let $V$ and $W$ be non-trivial normed vector spaces, then the operator norm of a
linear operator $T : V \to W$ is given by $\norm{T}_{\op} = \sup_{v : \norm{v} =
1} \norm{T v}$. When $V = \R^d$ and $W = \R^n$, the operator norm of a matrix
$A$ matches $\sigmax(A)$ and is also known as the spectral norm. We will drop
the sub-script in this case, meaning $\norm{A} = \norm{A}_{\op} = \sigmax(A)$.

For probability measures $\mu$ and $\nu$ on $\R^d$ with sufficiently many
moments, we denote by $\calW_q(\mu, \nu)$, $q \geq 1$ the transportation
distance \begin{align*}
  \calW_q\big(\mu, \nu\big) = \inf_\pi \left(\int \norm{\bfu - \bfv}_2^q\ \rmd
  \pi(\bfu, \bfv)\right)^{\tfrac{1}{q}}
\end{align*} where the infimum is taken over all probability measures $\pi$ on
$\R^d \times \R^d$ with marginals $\mu$ and $\nu$.

Given any two functions $f, g : \R^d \to \R$, $f(\bfx) = O\big(g(\bfx)\big)$ as
$\bfx \to \bfy$, signifies that $\limsup_{\bfx \to \bfy} \abs[\big]{f(\bfx) /
g(\bfx)} < \infty$. 

\section{Estimating Linear Predictors with Gradient Descent}
\label{sec::grad_desc}

Given a sample $(\bfX_i, Y_i)$, $i = 1, \ldots, n$ of data points $\bfX_i \in
\R^d$ with corresponding labels $Y_i \in \R$, we aim to learn a linear predictor
$\bfw \in \R^d$ such that $Y_i \approx \bfX_i\tran \bfw$ for each $i$. We will
focus on predictors learned via minimization of the empirical risk
\begin{align}
  \label{eq::lls_obj_erm}
  \weight \mapsto \dfrac{1}{n} \cdot \sum_{i = 1}^n \big(Y_i - \bfX_i\tran
  \weight\big)^2.
\end{align} Write $\bfY$ for the length $n$ vector with entries $Y_i$ and $X$
for the $(n \times d)$-matrix with rows $\bfX_i\tran$, then, up to the constant
factor $n\inv$, the empirical risk coincides with the linear least squares
objective \begin{align}
  \label{eq::lls_obj}
  \weight \mapsto \norm[\big]{\bfY - X \weight}_2^2.
\end{align} From this point on, we absorb the factor $n\inv$ into the observed
data and labels and work with the loss \eqref{eq::lls_obj}. To keep notation
compact, define the shorthand $\bbX = X\tran X$. If $\bbX\inv$ exists, then
\eqref{eq::lls_obj} admits the unique minimizer $\bbX\inv X\tran \bfY$, known as
the linear least squares estimator. The matrix $\bbX\inv X\tran$ defines a
pseudo-inverse for $X$, see \cite{campbell_meyer_2009}. For singular $\bbX$, the
definition of this estimator may be extended to take the value $X\pinv Y$, with
$X\pinv$ any generalization of the pseudo-inverse. In this case, $X$ admits many
different pseudo-inverses, which lead to distinct estimators. We will choose a
particular one so that $X\pinv \bfY$ coincides with the unique minimum-norm
minimizer of \eqref{eq::lls_obj}, see Appendix \ref{sec::svalue_pinv} for more
details.

\subsection{Noiseless Gradient Descent in the Overparametrized Setting}

We briefly discuss minimization of the linear least squares objective via
full-batch gradient descent. Fixing a sequence of step-sizes $\alpha_k > 0$ the
gradient descent recursion generated by \eqref{eq::lls_obj} takes the form
\begin{align*}
  \weight_{k + 1} &= \weight_k - \dfrac{\alpha_k}{2} \cdot
  \nabla_{\weight_k}\norm[\big]{\bfY - X \weight_k}_2^2,
\end{align*} started from some initial guess $\weight_1$. Expanding the norm in
\eqref{eq::lls_obj} yields the quadratic polynomial $\weight \mapsto
\weight\tran \bbX \weight - 2 \bfY\tran X \weight + \norm{\bfY}_2^2$, with
second derivative $\bbX$. Consequently, the objective is
$\sigmin(\bbX)$-strongly convex for invertible $\bbX$. Provided that $\alpha_k
\cdot \norm{\bbX} < 1$ for every $k \geq 1$ and $\sum_{k = 1}^\infty \alpha_k =
\infty$, this implies convergence of $\weight_k$ to the unique global minimizer
$\bbX\inv X\tran \bfY$, regardless of initialization. For singular $\bbX$, such
as in the over-parametrized regime with $d > n$, the objective
\eqref{eq::lls_obj} is merely convex, but an analogous convergence result holds
under additional assumptions. The gradient of \eqref{eq::lls_obj} evaluates to
$2 \cdot X\tran (\bfY - X \weight)$, so together with the property $\bbX X\pinv
= X\tran$ of the pseudo-inverse (Lemma \ref{lem::pinv_a}), we can rewrite the
gradient descent recursion as \begin{equation}
  \label{eq::rec_gd_nn}
  \begin{split}
    \weight_{k + 1} &= \big(I - \alpha_k \cdot \bbX\big) \weight_k - \alpha_k
    \cdot X\tran \bfY\\
    &= \big(I - \alpha_k \cdot \bbX\big) \big(\weight_k - X\pinv \bfY\big) +
    X\pinv \bfY.
  \end{split}
\end{equation} For a suitably chosen initial point $\weight_1$, the difference
$\weight_k - X\pinv \bfY$ always lies in a sub-space on which each matrix $I -
\alpha_k \cdot \bbX$ acts as a contraction, ensuring convergence of $\weight_k$
to $X\pinv \bfY$ as $k \to \infty$. A precise result is given in the following
classical lemma.

\begin{lemma}
  \label{lem::conv_gd}
  Suppose $\weight_1 \perp \ker(X)$ and $\sup_{\ell} \alpha_\ell \cdot
  \norm{\bbX} < 1$, then \begin{align*}
    \norm[\big]{\weight_{k + 1} - X\pinv \bfY}_2 \leq \exp \left(-
    \sigminp(\bbX) \cdot \sum_{\ell = 1}^k \alpha_\ell\right) \cdot
    \norm[\big]{\weight_1 - X\pinv \bfY}_2
  \end{align*} for every $k \geq 1$. Provided that $\sum_{\ell = 1}^\infty
  \alpha_\ell = \infty$, the left-hand side in the previous display vanishes as
  $k \to \infty$.
\end{lemma}

The fact that $\bfw_k$ converges to the minimizer $X\pinv \bfY$ of
\eqref{eq::lls_obj} with the smallest magnitude appears in Section 3 of
\cite{bartlett_montanari_et_al_2021} as an example of implicit regularization
through gradient descent. Lemma \ref{lem::conv_gd} restates this effect as an
explicit property of the initialization. Indeed, if $\bfw = \bfu + \bfv$ denotes
an orthogonal decomposition of $\bfw$ along the linear sub-space $\ker(X)$,
meaning $\bfu \perp \ker(X)$ and $\bfv \in \ker(X)$, then \begin{align}
  \label{eq::orth_rec}
  \big(I - \alpha_k \cdot \bbX\big) \bfw = \big(I - \alpha_k \cdot \bbX\big)
  \bfu + \bfv,
\end{align} so the recursion \eqref{eq::rec_gd_nn} always leaves the projection
of $\bfw_k$ onto $\ker(X)$ fixed. The assumption $\bfw_1 \perp \ker(X)$ requires
setting this projection to zero, which then gives the norm-minimal
initialization among all vectors with the same orthogonal component. The
gradient descent steps simply preserve this minimality, as shown in Lemma
\ref{lem::fixp_conv}. For a generic initialization, the iterates converge to the
minimum norm solution with the same orthogonal component as $\bfw_1$.

The convergence rate in Lemma \ref{lem::conv_gd} depends on the choice of
step-sizes $\alpha_k$. For example, a constant $\alpha_k = \alpha$ leads to
convergence at the rate $e^{-\alpha k}$, whereas linearly decaying step-sizes
$\alpha_k = \alpha / k$ yield the rate $k^{- \alpha}$ since $\sum_{\ell = 1}^k
\alpha / \ell \approx \alpha \cdot \log(k)$ for large enough $k$. In contrast,
generic convergence results for full-batch gradient descent on convex functions
only achieve the rate $k\inv$, even for constant step-sizes (Theorem 3.4 of
\cite{garrigos_gower_2024}).

\subsection{Random Weighting as a Generalization of Stochastic Gradient
Descent}

In practice, heavily over-parametrized models are often trained by evaluating
the gradient only on a subset of the available data during each iteration, due
to prohibitive cost of computing the full-batch gradient. This adds noise to the
gradient descent recursion and may be interpreted as a $\{0, 1\}$-valued random
weighting of the data points. Abstractly, given an initial guess $\whweight_1$
and step-sizes $\alpha_k > 0$, the randomly weighted gradient descent iterates
are defined as \begin{align}
  \label{eq::wgd_def}
  \whweight_{k + 1} = \whweight_k - \dfrac{\alpha_k}{2} \cdot
  \nabla_{\whweight_k} \norm[\big]{D_k \big(\bfY - X \whweight_k\big)}_2^2 =
  \big(I - \alpha_k \cdot X\tran D_k^2 X\big) \whweight_k + \alpha_k \cdot
  X\tran D_k^2 \bfY,
\end{align} with each $D_k$ an independent copy of a random $n \times n$
diagonal matrix $D$. During every iteration, this updates the parameter
$\whweight_k$ according to its fit on the weighted data $D_k X$ and $D_k \bfY$.
This definition includes various commonly used stochastic gradient descent
methods, but we postpone a discussion of specific examples until Section
\ref{sec::spec_sim}.

We shall not make any distributional assumptions on $D$, allowing for weightings
taking on any real values. For now, we only require that $D$ has finite moments
$M_p = \E[D^p]$ up to $p \leq 4$. As $D_k$ appears squared in
\eqref{eq::wgd_def}, this ensures that the iterates admit a well-defined
covariance matrix. Due to independence of the $D_k$, each randomized loss
$\norm{D_k \bfY - D_k X \weight}_2^2$ represents a Monte-Carlo estimate of
\begin{align}
  \label{eq::loss_exp}
  \E_D\Big[\norm[\big]{D \bfY - D X \weight}_2^2\Big] = \weight\tran X\tran
  \E\big[D^2\big] X \weight - 2 \bfY\tran \E\big[D^2\big] X \weight + \bfY\tran
  \E\big[D^2\big] \bfY = \norm[\Big]{\sqrt{M_2} \bfY - \sqrt{M_2} X
  \weight}_2^2,
\end{align} where $M^{1 / 2}$ denotes the unique positive semi-definite square
root of $M_2$. Multiplication by $D$ in \eqref{eq::loss_exp} may also be
understood as a random sketching with diagonal sketching matrix, see
\cite{vempala_2004}. However, the iterates \eqref{eq::loss_exp} do not minimize
a single randomly sketched loss $\norm{D \bfY - D X \weight}_2^2$, but rather
take a step along the gradient vector field of a newly sampled sketched loss in
each iteration, similar to iterative sketching methods such as
\cite{pilanci_wainwright_2016}.

For compactness sake, we will write $\whY = M_2^{1 / 2} \bfY$, $\whX = M_2^{1 /
2} X$, and $\whbbX = X\tran M_2 X$. The minimum-norm minimizer of
\eqref{eq::loss_exp} satisfies $\whweight = \whX\pinv \whY$, see Lemma
\ref{lem::pinv_c}. It then seems conceivable that $\whweight_k$ converges to
$\whweight$ in the sense that $\E_D\big[\norm{\whweight_k - \whweight}_2^2\big]$
vanishes as $k \to \infty$. Here, the expectation over $D$ refers to the
marginalization over the whole sequence of weighting matrices $D_1, D_2 \ldots$
that is sampled to generate the iterates $\whweight_k$. Indeed, the gradient
descent iterates may be expressed as \begin{align}
  \label{eq::gd_rw}
  \whweight_{k + 1} = \whweight_1 - \sum_{\ell = 1}^k \dfrac{\alpha_\ell}{2}
  \cdot \nabla_{\whweight_\ell} \norm[\big]{D_\ell \big(\bfY - X
  \whweight_\ell\big)}_2^2,
\end{align} suggesting that the contribution of the squared weighting matrices
$D_\ell^2$ could asymptotically average out to $M_2$ due to their independence.
Unfortunately, the situation is slightly more complicated as the summands in
\eqref{eq::gd_rw} are correlated across iterations. Further, in analogy with
\eqref{eq::rec_gd_nn}, we may rewrite the gradient descent recursion
\eqref{eq::wgd_def} as \begin{align}
  \label{eq::gd_var}
  \whweight_{k + 1} - \whweight = \big(I - \alpha_k \cdot X\tran D_k^2 X\big)
  \big(\whweight_k - \whweight\big) + \alpha_k \cdot X\tran D_k^2 \big(\bfY -
  X \whweight\big).
\end{align} This recursion represents the differences $\whweight_k -
\whweight$ as a vector auto-regressive (VAR) process with random coefficients,
mirroring the situation encountered in the study of linear regression with
dropout \cite{clara_langer_et_al_2024}. The random linear operators $\big(I -
\alpha_k \cdot X\tran D_k^2 X\big)$ and random shifts $X\tran D_k^2 \big(\bfY -
X \whweight\big)$ feature significant correlation, making it a priori unclear
whether a result analogous to Lemma \ref{lem::conv_gd} holds.

Whenever $M_2$ does not admit zero diagonal entries and $X$ has linearly
independent rows, Lemma \ref{lem::pinv_d} implies $\whX\pinv = X\pinv M^{- 1/
2}_2$ and so \begin{align}
  \label{eq::rows_indep}
  \bfY - X \whweight = \big(I - X \whX\pinv M_2^{1 / 2}\big) \bfY = \big(I - X
  X\pinv\big) \bfY = \bfzero,
\end{align} where the last equality follows from $I - X X\pinv$ being the
orthogonal projection onto the (trivial) kernel of $X\tran$. If $d > n$, then
$X$ having linearly independent rows equates to $\mathrm{rank}(X) = n$.
Accordingly, the affine shift in \eqref{eq::gd_var} vanishes, but this may not
necessarily be the case for correlated data. If \eqref{eq::rows_indep} is
satisfied, then \eqref{eq::gd_var} reduces to a linear recursion. As shown in
the analysis of simplified dropout in \cite{clara_langer_et_al_2024},
convergence of $\E\big[\norm{\whweight_k  - \whweight}_2^2\big]$ to zero follows
in that case.

To conclude this discussion, we briefly illustrate how the expression $X\tran
D_k^2 \big(\bfY - X \whweight\big)$ relates to the ``residual quantity at the
minimum'' that appeared in \cite{needell_srebro_et_al_2014}. For a generic
$\mu$-strongly convex loss $f(\bfx) = \E_D\big[f_i(\bfx)\big]$, where $i \sim D$
is the uniform distribution over finitely many individual losses $f_i$, this
quantity is defined as $\E_D\big[\norm{\nabla f_i(\bfx_*)}_2^2\big]$, with
$\bfx_*$ denoting the unique global minimum of $f$. The main result of
\cite{needell_srebro_et_al_2014} then shows that the stochastic gradient descent
iterates $\bfx_{k + 1} - \alpha \cdot \nabla f_{i_k}(\bfx_k)$, $i_k \sim D$
satisfy \begin{align}
  \label{eq::res_conv}
  \E_D\Big[\norm[\big]{\bfx_k - \bfx_*}_2^2\Big] \approx O\Big(e^{- \alpha \mu
  \cdot k}\Big) + O\Big(\alpha \cdot \E_D\big[\norm{\nabla
  f_i(\bfx_*)}_2^2\big]\Big),
\end{align} where we have omitted some constants related to the smoothness of
the $f_i$. Consequently, a non-zero residual acts as a fundamental lower-bound
to the estimation of $\bfx_*$ via constant step-size stochastic gradient
schemes. The setting of zero residual is also termed the realizable case. To
achieve a desired accuracy in squared norm, the second term in the previous
display must be controlled via the step-size $\alpha$. The gradient of $\norm{D
\bfY - D X \bfw}_2^2$ at $\whweight$ evaluates to $X\tran D^2 \big(\bfY - X
\whweight\big)$, so the second moment of the random shift in \eqref{eq::gd_var}
represents the residual quantity for the randomly weighted iterates
\eqref{eq::wgd_def}. Hence, whenever \eqref{eq::rows_indep} holds, the randomly
weighted linear regression problem is in the realizable case, meaning the
minimum-norm solution $\whweight$ features no residual variance. As a by-product
of our convergence analysis in the next section, we will obtain a precise
statement analogous to \eqref{eq::res_conv} for our specific regression model.

\section{Convergence Analysis for Generic Random Weightings}
\label{sec::conv_res_iso}

We will now focus on analyzing the convergence of the random dynamical system in
\eqref{eq::gd_var}, without making a specific distributional assumption on the
random weighting matrix $D$. As in the previous section, we use the shorthand
notations $\whY = M_2^{1 / 2} \bfY$, $\whX = M_2^{1 / 2} X$, $\whbbX = X\tran
M_2 X$, and $\whweight = \whX\pinv \whY$, as well as $\bbE_D$ for the
expectation with respect to the random matrices $D_k \sim D$, $k \geq 1$.

\subsection{Convergence of the First and Second Moments}

Our first goal is to assess convergence of the expectation and covariance of
$\whweight_k - \whweight$. We suppose the following standing assumptions hold,
which parallel the requirements of Lemma \ref{lem::conv_gd}.

\begin{assumption}
  \label{ass::conv_iso}
  \begin{enumerate}
    \item The step sizes satisfy $\sup_{\ell} \alpha_\ell \cdot \norm{\whbbX} <
      1$ and $\sum_{\ell = 1}^\infty \alpha_\ell = \infty$. \label{ass::conv_a}
    \item The initial guess $\whweight_1$ almost surely lies in the orthogonal
      complement of $\ker(X)$. \label{ass::conv_b}
    \item The random weighting matrix $D$ has finite fourth moment and $M_2 =
      \E\big[D^2\big]$ is non-singular. \label{ass::conv_c}
  \end{enumerate}
\end{assumption}

If $M_{2, ii} = 0$ for some entry $i$, then $D_{ii} = 0$ almost surely.
Accordingly, the corresponding data point $\bfX_i$ would never be active and can
be removed without changing the iterates \ref{eq::wgd_def}, making Assumption
\ref{ass::conv_c} natural. In analogy with \eqref{eq::orth_rec}, suppose $\bfw =
\bfu + \bfv$, where $\bfv$ and $\bfu$ are random vectors respectively
concentrated on $\ker(X)$ and its orthogonal complement, then \begin{align*}
  \big(I - \alpha_k \cdot X\tran D_k^2 X\big) \bfw = \big(I - \alpha_k \cdot
  X\tran D_k^2 X\big) \bfu + \bfv.
\end{align*} Consequently, the recursion \eqref{eq::gd_var} can never change the
orthogonal projection of $\whweight_k - \whweight$ onto $\ker(X)$. As in the
noiseless case, we will use this fact together with Assumption \ref{ass::conv_b}
to argue that $\whweight_k - \whweight$ always stays orthogonal to $\ker(X)$.

As in the analysis of linear regression with dropout
\cite{clara_langer_et_al_2024}, we start by marginalizing the contribution of
the weighting matrices $D_k$ to the evolution of the random dynamical system
\eqref{eq::gd_var}. By definition, $\E_D\big[I - \alpha_k \cdot X\tran D_k^2
X\big] = I - \alpha_k \cdot \whbbX$ and so independence of the weighting
matrices implies \begin{equation}
  \label{eq::exp_lin}
  \begin{split}
    \E_D\Big[\big(I - \alpha_k \cdot X\tran D_k^2 X\big) \big(\whweight_k -
    \whweight\big)\Big] &= \E_D\Bigg[\E_D\Big[\big(I - \alpha_k \cdot X\tran
    D_k^2 X\big) \big(\whweight_k - \whweight\big) \bigmid
    \whweight_k\Big]\Bigg]\\
    &= \big(I - \alpha_k \cdot \whbbX\big) \E_D\big[\whweight_k -
    \whweight\big].
  \end{split}
\end{equation} Consequently, the marginalized linear part of the affine
dynamical system \eqref{eq::gd_var} acts analogous to the noiseless recursion
\eqref{eq::wgd_def}, with the weighted matrix $\whbbX$ replacing $\bbX$.
Further, the random affine shift in \eqref{eq::gd_var} vanishes in mean since
\begin{align}
  \label{eq::exp_shift}
  \E_D\Big[X\tran D_k^2 \big(\bfY - X \whweight\big)\Big] = X\tran M_2 \big(\bfY
  - X \whweight\big) = \whX\tran \whY - \whbbX \whweight = \mathbf{0},
\end{align} where the last equality follows from Lemma \ref{lem::pinv_a} and
the definition of $\whweight$ together implying $\whbbX \whweight = \whbbX
\whX\pinv \whY = \whX\tran \whY$. Hence, marginalizing the algorithmic noise in
\eqref{eq::gd_var} yields a linear dynamical system that satisfies a
convergence result paralleling Lemma \ref{lem::conv_gd}.

\begin{lemma}
  \label{lem::conv_exp_iso}
  Under Assumption \ref{ass::conv_iso}, for every $k \geq 1$
  \begin{align*}
    \norm[\Big]{\E_D\big[\whweight_{k + 1} - \whweight\big]}_2 \leq \exp \left(-
    \sigminp(\whbbX) \cdot \sum_{\ell = 1}^k \alpha_\ell\right) \cdot
    \norm[\big]{\whweight_1 - \whweight}_2.
  \end{align*}
\end{lemma}

Comparing with Lemma \ref{lem::conv_gd}, the convergence rate of the randomly
weighted iterates depends on $\sigminp(\whbbX)$, as opposed to $\sigminp(\bbX)$.
If $\sigminp(\whbbX) > \sigminp(\whX)$, this leads to faster convergence, which
may also be interpreted as an improvement of the effective conditioning of the
problem in expectation.

With the marginalized dynamics taken care of in Lemma \ref{lem::conv_exp_iso},
we move on to assess how the differences $\whweight_k - \whweight$ diffuse
around their expectations $\E_D\big[\whweight_k - \whweight\big]$ due to the
random weighting $D_k$. Since the dynamics of $\whweight_k - \whweight$ admit
the affine representation \eqref{eq::gd_var}, we may expect to discover an
affine structure in the evolution of the second moments $\E_D\big[(\whweight_k -
\whweight) (\whweight_k - \whweight)\tran\big]$. This turns out to be almost
true, up to a vanishing remainder term. We recall that $A \odot B$ denotes the
element-wise product of two matrices.

\begin{lemma}
  \label{lem::var_op_iso}
  Write $\Sigma_D$ for the covariance matrix of the random vector
  $\big(D_{11}^2, \ldots, D_{nn}^2\big)$, well-defined due to Assumption
  \ref{ass::conv_c}, and consider the parametrized family $S_\alpha(\ \cdot\ )$,
  $\alpha > 0$ of affine operators acting on $(d \times d)$-matrices via
  \begin{align*}
    S_\alpha(A) = \big(I - \alpha \cdot \whbbX\big) A \big(I - \alpha \cdot
    \whbbX\big) + \alpha^2 \cdot X\tran \Bigg(\Sigma_D \odot \bigg(X A X\tran +
    \big(\bfY - X \whweight\big) \big(\bfY - X \whweight\big)\tran\bigg)\Bigg)
    X.
  \end{align*} Under Assumption \ref{ass::conv_b}, for every $k \geq 1$, there
  exists a symmetric $(d \times d)$-matrix $\rho_k$ such that $\ker(X) \subset
  \ker(\rho_k)$ and \begin{align*}
    \E_D\Big[\big(\whweight_{k + 1} - \whweight\big) \big(\whweight_{k + 1} -
    \whweight\big)\tran\Big] - S_{\alpha_k}\Bigg(\E_D\Big[\big(\whweight_k -
    \whweight\big) \big(\whweight_k - \whweight\big)\tran\Big]\Bigg) = \rho_k,
  \end{align*} If in addition Assumption \ref{ass::conv_a} holds, then for $k >
  1$ the remainder term vanishes at the rate \begin{align*}
    \norm[\big]{\rho_k} \leq 2 \alpha_k^2 \cdot \norm{X}^3 \cdot \norm{\Sigma_D}
    \cdot \exp \left(- \sigminp(\whbbX) \cdot \sum_{\ell = 1}^{k - 1}
    \alpha_\ell\right) \cdot \norm[\big]{\whweight_1 - \whweight}_2 \cdot
    \norm[\big]{\bfY - X \whweight}_2.
  \end{align*}
\end{lemma} Lemma \ref{lem::var_op_iso} may be summarized as follows: up to an
exponentially small remainder term, the second moment of $\whweight_{k + 1} -
\whweight$ evolves as an affine dynamical system, pushed forward by the
time-dependent maps $S_{\alpha_k}$. For constant step-sizes $\alpha_k = \alpha$,
the iteration map $S_\alpha$ stays unchanged in time. As a shorthand, we will
use $\Sint_{\alpha_k} = S_{\alpha_k}(0)$ and $\Slin_{\alpha_k}(\ \cdot\ ) =
S_{\alpha_k}(\ \cdot\ )- \Sint_{\alpha_k}$ to refer to the intercept and linear
part of each affine map $S_{\alpha_k}$. Both $S_{\alpha_k}$ and $\rho_k$ may be
computed directly from the VAR representation \eqref{eq::gd_var}. In abstract
terms, \eqref{eq::gd_var} takes the form $\bfz_{k + 1} = G_k(\bfz_k) +
\bm{\xi}_k$, with $G_k$ and $\bm{\xi}_k$ sequences of independent random linear
operators and affine shifts, meaning \begin{align*}
  \bfz_{k + 1} \bfz_{k + 1}\tran = G_k(\bfz_k) G_k(\bfz_k)\tran + \bm{\xi}_k
  \bm{\xi}_k\tran + G_k(\bfz_k) \bm{\xi}_k\tran + \bm{\xi}_k G_k(\bfz_k)\tran.
\end{align*} The linear operator $\Slin_{\alpha_k}$ then corresponds to the
second moment of $G_{k}(\bfz_k)$ and the intercept $\Sint_{\alpha_k}$ to the
second moment of $\bm{\xi}_k$. The remainder $\rho_k$ consists of the
cross-multiplied terms, which can be bounded via Lemma \ref{lem::conv_exp_iso}.
As shown in \eqref{eq::rows_indep}, $\bm{\xi}_k = \bfzero$ almost surely
whenever $X$ features independent rows. In this case, $\Sint_{\alpha_k} = \rho_k
= 0$ and so the dynamic in Lemma \ref{lem::var_op_iso} collapses to an exact
linear evolution. Using a similar technique as in the analysis of simplified
dropout in \cite{clara_langer_et_al_2025}, linearity of the recursion implies
vanishing of the second moment for small enough step-sizes, so we will mainly
focus on the case of linearly dependent rows.

We may now proceed to derive a limiting expression and a convergence rate for
the second moment of $\whweight_k - \whweight$ by unraveling the recursion in
Lemma \ref{lem::var_op_iso}. The affine maps $S_{\alpha_k}$ depend on the
sequence of step-sizes, so the limit will also depend on this choice. For the
sake of simplicity, we present limits for two classical approaches:
square-summable and constant step-sizes. As an intermediate step in proving the
next result, a limiting expression for generic step-sizes is given in Lemma
\ref{lem::var_ass}.

\begin{theorem}
  \label{thm::var_conv_iso}
  In addition to Assumption \ref{ass::conv_iso}, suppose \begin{align*}
    \sup_{\ell} \alpha_\ell < \dfrac{\sigminp(\whbbX)}{\sigminp(\whbbX)^2 +
    \norm{X}^4 \cdot \norm{\Sigma_D}},
  \end{align*} then the following hold: \begin{enumerate}
  \item If $\sum_{\ell = 1}^\infty \alpha_\ell^2 < \infty$, then the second
    moment of $\whweight_{k + 1} - \whweight$ vanishes as $k \to \infty$. In
    particular, if $\alpha_k = \alpha / k$ for some $\alpha > 0$, then there
    exists a constant $C_1$ that depends on $X$, $Y$, $\whweight_1$, $\alpha$
    and the first four moments of $D$, such that \begin{align*}
        \norm[\Bigg]{\E_D\Big[\big(\whweight_{k + 1} - \whweight\big)
        \big(\whweight_{k + 1} - \whweight\big)\tran\Big]} \leq C_1 \cdot
        \dfrac{1}{k^{\alpha \sigminp(\whbbX)}}.
      \end{align*} \label{thm::var_conv_iso_a}
    \item If $\alpha_k = \alpha$ for every $k$, then there exists a finite
      constant $C_2$ that depends on $X$, $Y$, $\whweight_1$, $\alpha$, and the
      first four moments of $D$, such that \begin{align*}
        \norm[\Bigg]{\E_D\Big[\big(\whweight_{k + 1} - \whweight\big)
        \big(\whweight_{k + 1} - \whweight\big)\tran\Big] - \big(\id -
        \Slin_\alpha\big)\inv_{\ker(X)} \big(\Sint_\alpha\big)} \leq C_2 \cdot
        \big(2 + k \alpha^2\big) \cdot \exp\Big(- \alpha \sigminp(\whbbX) \cdot
        \big(k - 1\big)\Big),
      \end{align*} where $(\ \cdot\ )_{\ker(X)}\inv$ refers to inversion on the
      subspace of matrices $A$ satisfying $\ker(X) \subset \ker(A)$.
      \label{thm::var_conv_iso_b}
  \end{enumerate}
\end{theorem}

The operator $\id - \Slin_\alpha$ in Theorem \ref{thm::var_conv_iso_b} may not
admit a global inverse whenever $\whbbX$ has a non-trivial kernel. One can show
that $\Slin_\alpha$ acts as a contraction on the subspace of matrices with
kernel including $\ker(X)$, hence the resulting Neumann series converges to the
inverse of $\id - \Slin_\alpha$ on this subspace. This also shows necessity of
Assumption \ref{ass::conv_b}, without it $\Slin_\alpha$ may not contract the
norm of $(\whweight_1 - \whweight) (\whweight_1 - \whweight)\tran$ through each
repeated application, preventing convergence to the claimed limit.

For an explicit statement of the constants $C_1$ and $C_2$, see the proof of
Theorem \ref{thm::var_conv_iso}. Square summability of the step-sizes is also
known as the Robbins-Monro condition, after the seminal work
\cite{robbins_monro_1951}. The randomized gradient in \eqref{eq::gd_rw} always
appears scaled by $\alpha_k / 2$, so its variance admits control via
$\alpha_k^2$. Square summability then ensures that the randomized gradients
converge to their expectations sufficiently fast as $k \to \infty$. In this
case, the (deterministic) vector field of the expected gradients asymptotically
dominates the dynamics, which yields convergence to a critical point of the
underlying loss in quite general settings \cite{davis_drusvyatskiy_et_al_2020,
dereich_kassing_2024}. In our setting, the only critical point reachable from an
initialization $\whweight_1 \perp \ker(\whX)$ is $\whweight$. Convergence in
spectral norm, as in Theorem \ref{thm::var_conv_iso_a} also entails convergence
in second mean since \begin{align*}
  \E_D\Big[\norm[\big]{\whweight_k - \whweight}_2^2\Big] =
  \Tr\Bigg(\E_D\Big[\big(\whweight_{k + 1} - \whweight\big) \big(\whweight_{k + 1}
  - \whweight\big)\tran\Big]\Bigg).
\end{align*} For constant step-sizes, Theorem \ref{thm::var_conv_iso_b} shows
that the latter trace cannot vanish, unless the matrix $(\id -
\Slin_\alpha)\inv_{\ker(X)} (\Sint_\alpha)$ is nilpotent. This is the case
whenever \eqref{eq::rows_indep} holds, which implies $\Sint_{\alpha} = 0$. As
discussed in the final paragraph of Section \ref{sec::grad_desc}, this relates
to the residual quantity at the minimum in \cite{needell_srebro_et_al_2014}.
Taking the trace yields a result analogous to \eqref{eq::res_conv}, with the
slightly slower rate $O\big(k \cdot e^{- k}\big)$. This is due to Theorem
\ref{thm::var_conv_iso_b} proving a stronger form of convergence via the
spectral norm, whereas \cite{needell_srebro_et_al_2014} exploit the
co-coercivity of smooth functions to receive the rate $O\big(e^{- k}\big)$,
which only holds in the squared norm $\E_D\big[\norm{\whweight_k -
\whweight}_2^2\big]$. An equivalent of \eqref{eq::res_conv} with the faster rate
will follow from the results in the next section.

\subsection{Geometric Moment Contraction and Stationary Distributions}

Together, Lemma \ref{lem::conv_exp_iso} and Theorem \ref{thm::var_conv_iso} show
that the first two moments of $\whweight_k$ converges as $k \to \infty$, but
this does not by itself imply the existence of a well-defined limit for the
distribution of $\whweight_k$. Under square-summable step-sizes, Theorem
\ref{thm::var_conv_iso_b} yields convergence to a point mass. For constant
step-sizes, stochastic gradient descent may admit non-degenerate long-run
distributions that reflect the local geometry of the underlying loss function
\cite{azizian_iutzeler_et_al_2024}.

We will analyze the long-run distribution of the random dynamical system
\eqref{eq::gd_var} by adapting the techniques used in
\cite{li_schmidt-hieber_et_al_2024}. Fix two suitable measures $\mu$ and $\nu$
on $\R^d$. We now define the coupled stochastic processes $\whu_k$ and $\whv_k$,
initialized by $\whu_1 \sim \mu$ and $\whv_1 \sim \nu$, that each obey the
recursion \eqref{eq::gd_rw} with exactly the same sequence of outcomes $D_1,
D_2, \ldots$ for the random weighting matrices, sampled independent of the
initializations. Following \cite{li_schmidt-hieber_et_al_2024}, the gradient
descent iterates are said to satisfy geometric moment contraction (GMC) if
\begin{align}
  \label{eq::gmc}
  \E_D\Big[\norm[\big]{\whu_{k + 1} - \whv_{k + 1}}_2^q\Big]^{1 / q} = r_q^k
  \cdot \norm[\big]{\whu_1 - \whv_1}_2^q
\end{align} for some constant $r_q \in (0, 1)$ and all $q \geq 1$ that admit a
finite left-hand side expectation. The GMC property is a key tool in the
analysis of iterated random functions, for example by ensuring the existence of
unique stationary distributions, see \cite{wu_shao_2004} for further details.

We emphasize that \eqref{eq::gmc} only concerns the algorithmic randomness,
without reference to the initial distribution, or any further randomness
inherent to the data $(X, \bfY)$. It is possible to account for the latter, see
\cite{li_schmidt-hieber_et_al_2024}, but we are mainly interested in the
algorithmic randomness, so we do not model the data generation process further.
For $r_q$ independent of the initial distributions $\mu$ and $\nu$, we may
integrate over the product measure $\mu \otimes \nu$ on both sides of
\eqref{eq::gmc}, which by Fubini's Theorem yields geometric moment contraction
with respect to the joint distribution of $\whu_1$, $\whv_1$, and $D$. To prove
the GMC property, we will use the following additonal assumptions:

\begin{assumption}
  \label{ass::gmc}
  \begin{enumerate}
    \item The distribution of $D$ has compact support, meaning $\norm{D} \leq
      \tau$ almost surely for some $\tau < \infty$. \label{ass::gmc_b}
    \item The algorithm \eqref{eq::gd_rw} is run with constant step-sizes
      $\alpha_k = \alpha$ that satisfy $\alpha \tau^2 \cdot \norm{\bbX} < 2$.
      \label{ass::gmc_c}
  \end{enumerate}
\end{assumption}

Adapting the argument that leads to Lemma 1 in
\cite{li_schmidt-hieber_et_al_2024} now yields the GMC property for the randomly
weighted gradient descent iterates. In particular, convergence towards a unique
stationary distribution may be phrased in terms of the transportation distances
$\calW_q$, $q \geq 1$ as defined in Section \ref{sec::not}. We recall that
$\calW_q$ metrizes both weak convergence and convergence of the first $q$
moments, see Chapter 7 of \cite{villani_2003} for more details.

\begin{theorem}
  \label{thm::stat_dist}
  Suppose Assumption \ref{ass::conv_iso} and \ref{ass::gmc} both hold, then the
  gradient descent iterates satisfy the GMC property \eqref{eq::gmc} with
  \begin{align*}
    r_q^q \leq 1 - \alpha \cdot \big(2 - \alpha \tau^2 \cdot \norm{\bbX}\big)
    \cdot \sigminp(\whbbX) < 1.
  \end{align*} Consequently, the iterates \eqref{eq::gd_rw} admit a unique
  stationary distribution that does not depend on the initialization. If
  $\widehat{\mu}_k$ and $\widehat{\mu}_\infty$ respectively denote the measures
  induced by $\whweight_k$ and a random vector following the stationary
  distribution, then \begin{align*}
    \calW_q\big(\widehat{\mu}_k, \widehat{\mu}_\infty\big) \leq C_3 \cdot
    \exp\left(- \dfrac{\alpha \cdot \big(2 - \alpha \tau^2 \cdot
    \norm{\bbX}\big) \cdot \sigminp(\whbbX)}{q} \cdot k\right)
  \end{align*} for all $q \geq 1$, with constant $C_3 > 0$ depending only on
  $X$, $\alpha$, $\tau$, and the distribution of $\whweight_1$.
\end{theorem}

The stationary distribution in the previous theorem may be visualized as
follows. Suppose $D_{j} \sim D$ for all $j \in \Z$ and define $\whweight_\infty$
such that \begin{align}
  \label{eq::stat_vec}
  \whweight_\infty - \whweight = \alpha \cdot \sum_{j = 0}^{- \infty}\
  \left(\prod_{i = 0}^{j + 1} \big(I - \alpha \cdot X\tran D_j^2 X\big)\right)
  X\tran D_j^2 \big(\bfY - X \whweight\big),
\end{align} with the convention that the product evaluates to the identity
matrix whenever $i \leq j$. By construction, performing $k$ steps of the
iteration \eqref{eq::gd_var} with initialization $\whweight_\infty - \whweight$
simply shifts the indexes $i$ and $j$ to now start at $k$. Since the random
weighting matrices are i.i.d., this leaves the distribution of $\whweight_\infty
- \whweight$ fixed, so the latter may be seen as an independent copy of the
stationary distribution. In contrast, the distribution reached by the iterates
$\whweight_k$ only depends on $D_k$, $k \geq 1$ as shown in \cite{wu_shao_2004}.

If \eqref{eq::rows_indep} holds, then $\widehat{\mu}_\infty$ constitutes a point
mass at $\whweight$ since the right-hand side of \eqref{eq::stat_vec} equals
$\bfzero$ almost surely. As discussed at the end of Section
\ref{sec::grad_desc}, this corresponds to the realizable case in
\cite{needell_srebro_et_al_2014}, when the gradient of the loss features no
residual variance at the minimum. A slightly weaker result may be proven in the
non-realizable case by controlling the magnitude of the residual variance via
the step-size $\alpha$, see also Corollary 2.2 of
\cite{needell_srebro_et_al_2014}. Using the convergence towards a unique
stationary distribution in Theorem \ref{thm::stat_dist}, we can derive an
analogous result that bounds the distance of $\widehat{\mu}_k$ to a point mass
at $\whweight$ up to a given tolerance level.

\begin{theorem}
  \label{thm::conv_point}
  Suppose that requirements of both Theorem \ref{thm::var_conv_iso_b} and
  Theorem \ref{thm::stat_dist} hold, as well as $\bfY - X \whweight \neq
  \bfzero$. Fix $\eps > 0$ and assume additionally that \begin{align*}
    \alpha \leq \dfrac{\sigminp(\whbbX) \cdot \eps^2}{d \cdot \norm{\Sigma_D}
    \cdot \norm{\bbX} \cdot \norm[\big]{\bfY - X \whweight}_2}.
  \end{align*} Then, for all \begin{align*}
    k > \dfrac{2 \log\big(2 C_3 / \eps\big)}{\alpha \cdot \big(2 - \alpha
    \tau^2 \cdot \norm{\bbX}\big) \cdot \sigminp(\whbbX)},
  \end{align*} the distribution of the randomly weighted iterates
  \eqref{eq::gd_rw} satisfies $\calW_2\big(\widehat{\mu}_k,
  \delta_{\whweight}\big) < \eps$, with $\delta_{\whweight}$ denoting a point
  mass at $\whweight$.
\end{theorem}

The proof of the previous theorem proceeds along the following lines: the
assumption on $k$ controls the distance towards the stationary distribution
$\widehat{\mu}_\infty$ via Theorem \ref{thm::stat_dist}, then a bound for
$\calW_2\big(\widehat{\mu}_\infty, \delta_{\whweight}\big)$ in terms of $\alpha$
enables reverse engineering of the admissible step-sizes. The interrelated
quantities $\eps$, $\alpha$, and $k$ together determine the speed of
convergence. A smaller desired tolerance level $\eps$ demands shrinking of the
step-size $\alpha$, which in turn requires a larger number of iterations $k$ to
converge within the prescribed tolerance. In comparison with Corollaries 2.1 and
3.1 of \cite{needell_srebro_et_al_2014}, Theorem \ref{thm::stat_dist} only
demands logarithmic dependence of $k$ on $\eps\inv$.

\section{Properties of Specific Random Weightings}
\label{sec::spec_sim}

Having characterized the effect of random weightings on the gradient descent
dynamics, we move on to discuss some specific sampling strategies for the
coefficients $D_{k, ii}$ in \eqref{eq::wgd_def}. We will focus on $\{0,
1\}$-valued outcomes as the most important case, while keeping in mind that much
of the discussion in this section also applies to weightings with continuous
outcomes. Abstractly, any binary weighting may be interpreted as a distribution
over subsets $S$ of $\{1, \ldots n\}$, where $D_{k, ii} = 1$ if $i \in S$.
Sampling $D_k X$ and $D_k \bfY$ then selects the mini-batch corresponding to the
subset $S$ from the available data on which to perform the gradient update in
\eqref{eq::wgd_def}. The classic instance of stochastic gradient descent (SGD)
samples each $D_k$ with a single diagonal entry equaling $1$ with uniform
probability $1 / n$, hence selecting a single data point per iteration.

As shown in \cite{needell_srebro_et_al_2014}, non-uniform sampling can
accelerate the convergence of SGD by prioritizing data points that are
especially influential to the overall loss \eqref{eq::lls_obj_erm}. For each $i
= 1, \ldots, n$ \begin{align*}
  \dfrac{1}{2} \cdot \norm[\Big]{\nabla_{\bfw} \big(Y_i - \bfX_i\tran
  \bfw\big)^2 - \nabla_{\bfv} \big(Y_i - \bfX_i\tran \bfv\big)^2}_2 =
  \norm[\big]{\bfX_i\tran (\bfw - \bfv)}_2,
\end{align*} meaning any change in parameter $\bfw$ can move towards a critical
point of $(Y_i - \bfX_i\tran \bfw)^2$ with size roughly proportional to
$\norm{\bfX_i}_2$. In other words, the norm of $\bfX_i$ gives an estimate of the
sensitivity of the loss to a change in fit $Y_i - \bfX_i\tran \bfw$ on the
$i$\textsuperscript{th} data point. Running weighted SGD with inclusion
probabilities $\bbP\big(D_{k, ii} = 1\big) \propto \norm{\bfX_i}_2$ leads to the
randomized Kaczmarz method of \cite{strohmer_vershynin_2009}, as already noted
in Section 5 of \cite{needell_srebro_et_al_2014}. This method achieves
exponential convergence on full-rank over-parametrized linear least squares
problems, which may be interpreted as a specific case of Theorem
\ref{thm::stat_dist}. In the full-rank case, the stationary distribution
$\widehat{\mu}_{\infty}$ in Theorem \ref{thm::stat_dist} collapses to the point
mass $\delta_{\whweight}$, centered at the weighted linear least squares
estimator. Applying Lemma \ref{lem::pinv_d}, the latter coincides with the usual
linear least squares estimator $X\pinv \bfY$ in the full-rank setting, so the
desired convergence rate follows.

In general, any element of the probability simplex $\big\{\bfp \in \R^n \mid p_i
\geq 0 \mbox{ and } \sum_{i = 1}^n p_i = 1\big\}$ corresponds to a valid $\{0,
1\}$-valued weighted sampling scheme. As shown in Section
\ref{sec::conv_res_iso}, the chosen probability vector $\bfp$ affects the
weighted SGD algorithm both through the expected squared weighting matrix $M_2 =
\E\big[D^2\big]$ and the covariance matrix $\Sigma_D$ of the random vector
$\big(D_{11}^2, \ldots, D_{nn}^2\big)$. Binary outcomes entail $D^2 = D$,
meaning \begin{align*}
  M_{2, ii} &= \E\big[D_{ii}\big] = p_i\\
  \Sigma_{D, ij} &= \Cov\big(D_{ii}, D_{jj}\big) = \E\Big[\big(D_{ii} - p_i\big)
  \big(D_{jj} - p_j\big)\Big]\\
  &= \E\big[D_{ii} D_{jj}\big] - p_i \cdot \E\big[D_{jj}\big] -
  \E\big[D_{ii}\big] \cdot p_j + p_i p_j = \begin{cases}
    p_i (1 - p_i) &\mbox{ if } i = j\\
    - p_i p_j &\mbox{ otherwise,}
  \end{cases}
\end{align*} where the last equality follows from $D_{ii} = 1$ implying $D_{jj}
= 0$ for batches of size $1$. In particular, $\Sigma_D$ may be written in the
form $M_2 - \bfp \bfp\tran$.

\begin{figure}[tb]
    \centering
    \includegraphics[width = 0.75\textwidth]{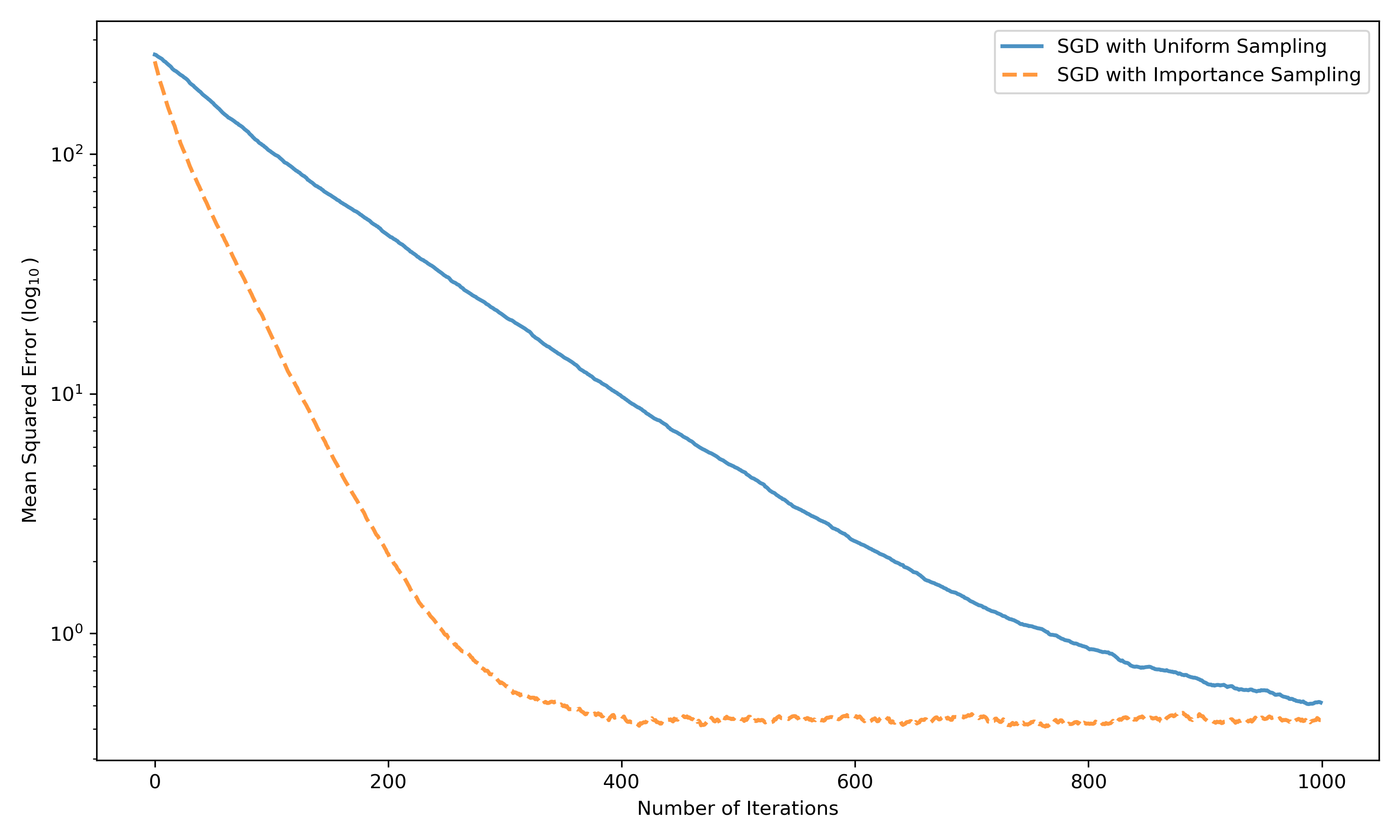}
    \caption{Convergence in squared distance $\E\big[\norm{\whweight_k -
    \whweight}_2^2\big]$ of SGD with uniform sampling, $p_i = 1 / n$, and
    importance sampling, $p_i = \exp\big(\norm{\bfX_i}_2\big) / \sum_{i = 1}^n
    \exp\big(\norm{\bfX_i}_2\big)$. Both algorithms are run on the same
    synthetic data set with normally distributed entries and the same constant
    step-size $\alpha$. To create a setting where importance sampling yields a
    noticeable benefit, a random subset of the data points $\bfX_i$ have been
    rescaled to have larger norm than the rest.}
    \label{fig::speed_comparison}
\end{figure}

For a constant step-size $\alpha_k = \alpha > 0$, the convergence rates
presented in Section \ref{sec::conv_res_iso} scale exponentially with $\alpha
\cdot \sigminp(\whbbX)$, where $\whbbX = X\tran M_2 X$. Assumption
\ref{ass::conv_a} requires $\alpha \cdot \norm{\whbbX} < 1$ and in turn the
effective inverse condition number $\sigminp(\whbbX) / \norm{\whbbX}$ of the
weighted matrix $\whbbX$ bounds the exponent. Hence, the largest and smallest
inclusion probabilities determine the achievable speed-up via \begin{align*}
  \dfrac{\sigminp(\whbbX)}{\norm{\whbbX}} \leq \dfrac{\max_{i = 1, \ldots, n}
  p_i}{\min_{i = 1, \ldots, n} p_i} \cdot \dfrac{\sigminp(\bbX)}{\norm{\bbX}}.
\end{align*} See Figure \ref{fig::speed_comparison} for an empirical
illustration of the speed-up. In fact, $D$ only affects the convergence rate
through $M_2$, meaning the maximal increase in exponent always depends on the
ratio between $\max_{i = 1, \ldots, n} \E\big[D_{ii}^2\big]$ and $\min_{i = 1,
\ldots, n} \E\big[D_{ii}^2\big]$, even for non-binary weighting distributions.
The actual gain may, however, be lower. For example, if $X$ has rank $1$, then
$\sigminp(\whbbX) / \norm{\whbbX} = \sigminp(\bbX) / \norm{\bbX} = 1$ and no
speed-up occurs.

As shown in Theorem \ref{thm::var_conv_iso_b}, the affine operator $S_{\alpha}$
of Lemma \ref{lem::var_op_iso} describes how the iterates diffuse around their
expected path towards $\whweight$. Repeating the argument in \eqref{eq::rem_est}
without the trace operator leads to the estimate \begin{align}
  \norm[\Big]{\big(\id - \Slin_{\alpha}\big)\inv_{\ker(X)}
  \big(\Sint_{\alpha}\big)} \leq \dfrac{\alpha \cdot \norm{\bbX} \cdot
  \norm{\Sigma_D}}{\sigminp(\whbbX)} \cdot \norm[\big]{\bfY - X \whweight}_2^2
  &\leq \dfrac{\norm{X}^2 \cdot \norm{\Sigma_D}}{\sigminp(\whbbX)^2 + \norm{X}^4
  \cdot \norm{\Sigma_D}} \cdot \norm[\big]{\bfY - X \whweight}_2^2 \nonumber\\
  &\leq \dfrac{1}{\norm{X}^2} \cdot \norm[\big]{\bfY - X \whweight}_2^2
  \label{eq::var_bound_wfit}
\end{align} where the second inequality follows from the assumption on $\alpha$
in Theorem \ref{thm::var_conv_iso}. Rescaling such that $\norm{X} = 1$, the
previous display remains bounded by $\norm{\bfY - X \whweight}_2^2$. This may
seem surprising; $\Sigma_D$ measures the variance added by the random weighting,
so one may expect an increase in $\norm{\Sigma_D}$ to cause a larger long-run
variance of the iterates. However, the condition on $\alpha$ in Theorem
\ref{thm::var_conv_iso} automatically regulates the variance level.

This further underlines the importance of designing a weighting scheme that
interacts well with the linear regression problem \eqref{eq::lls_obj}, to
prevent a ``bad'' weighted solution $\whweight$. The iterates asymptotically
cluster near $\whweight$, so the fit of the latter holds great influence over
the long-run statistical properties of the algorithm, as illustrated by the next
result.

\begin{theorem}
  \label{thm::asym_srisk}
  Suppose $\bfY = X \starweight + \bfeps$ for some true unknown parameter
  $\starweight \in \R^d$ and independent centered measurement noise $\bfeps$
  with covariance matrix $\Sigma_{\bfeps}$. If $\norm{X} = 1$ and $\alpha_\ell =
  \alpha > 0$ satisfies the assumptions of Theorem \ref{thm::var_conv_iso_b},
  then \begin{align*}
    C\big(X, \starweight, M_2\big) \leq \lim_{k \to \infty}
    \E\Big[\norm[\big]{\starweight - \whweight_k}_2^2\Big] \leq C\big(X,
    \starweight, M_2\big) + \Tr\Big(\big(I - X \whX\pinv M_2^{1 / 2}\big)
    \Sigma_{\bfeps} \big(I - X \whX\pinv M_2^{1 / 2}\big)\tran\Big)
  \end{align*} for any deterministic initialization $\whweight_1$, where
  \begin{align*}
    C\big(X, \starweight, M_2\big) = \E\Big[\norm[\big]{\starweight -
    \whweight}_2^2\Big] = \norm[\Big]{\big(I - X\pinv X\big) \starweight}_2^2 +
    \Tr\Big(\big(\whX\pinv M_2^{1 / 2}\big) \Sigma_{\bfeps} \big(\whX\pinv
    M_2^{1 / 2}\big)\tran\Big),
  \end{align*} with expectation taken over both $\bfeps$ and the
  sequence of random weighting matrices $D_k$, $k \geq 1$.
\end{theorem}

The result shows that asymptotic recovery of $\starweight$ depends almost
entirely on the quality of $\whweight$ as an estimator of $\starweight$. This
may seem unexpected; different weighting distributions yield roughly comparable
performance in the asymptotic regime if they feature the same expected weighting
matrix $M_2$, regardless of the actual shape of the distribution. For example,
given a probability vector $\bfp$, any binary sampling scheme such that
$\bbP\big(D_{k, ii} = 1\big) = p_i$ yields similar asymptotic performance since
$M_{2, ii} = \E\big[D_{k, ii}\big] = p_i$ for any such distribution. This
includes weighted SGD \cite{needell_srebro_et_al_2014}, weighted mini-batch SGD
\cite{needell_ward_2017, csiba_richtarik_2018}, and independent sampling $D_{k,
ii} \sim \Ber(p_i)$ as in the stochastic Metropolis-Hastings method analyzed in
\cite{bieringer_kasieczka_et_al_2023}. Further, $D_{k, ii} \sim
\mathrm{Laplace}(p_i - \sigma^2, \sigma^2)$ gives an example of a heavy-tailed
weighting with continuous outcomes for which the iterates will cluster near the
same weighted estimator $\whweight$.

Intuitively, this phenomenon appears due to a Gauss-Markov like property of the
weighted estimator $\whweight$, in combination with the specific form of the
affine operator $S_\alpha$ in Lemma \ref{lem::var_op_iso}. By definition,
$\whweight$ is a deterministic linear transformation of $\bfY$, whereas each
iterate $\whweight_k$ results from a random affine transformation of $\bfY$.
Theorem 3 of \cite{clara_langer_et_al_2024} then shows that the covariance
between $\whweight$ and $\whweight_k - \whweight$ roughly scales with
$\E\big[\whweight_k - \whweight\big]$. Convergence in expectation (Lemma
\ref{lem::conv_gd}) now implies that the two are asymptotically uncorrelated, so
$\whweight_\infty$ must always have at least as much variance as $\whweight$
while featuring the same bias. This yields the lower bound in Theorem
\ref{thm::asym_srisk}, with the upper bound resulting from an estimate for the
additional variance in terms of $\bfY - X \whweight$, analogous to the
calculation \eqref{eq::var_bound_wfit}.

The lower bound in Theorem \ref{thm::asym_srisk} represents the fundamental
bias-variance trade off in recovering $\starweight$ via the weighted least
squares estimator $\whweight$, with $\big(I - X\pinv X\big) \starweight$ giving
the bias term and the trace expression giving the variance term. One may expect
that the bias should read $\big(I - \whX\pinv \whX\big) \starweight$, but this
turns out to be equivalent as $I - A\pinv A$ gives the orthogonal projection
onto $\ker(A)$ and $\ker(X) = \ker(\whX)$. If $\whX\pinv = (M^{1 / 2}_2 X)\pinv
= X\pinv M_2^{- 1 / 2}$, the variance term reduces to $\Tr\big(X\pinv
\Sigma_{\eps} (X\pinv)\tran\big)$ and $C(X, \alpha, M_2)$ matches the risk of
the usual linear least squares estimator $X\pinv \bfY$. The identity $(A B)\pinv
\neq B\pinv A\pinv$ only holds under specific conditions, see Lemma
\ref{lem::pinv_d} for a sufficient one and Theorem 1.4.1 in
\cite{campbell_meyer_2009} for a more general result.

In fact, a ``bad'' expected weighting matrix $M_2$ has the potential to create
arbitrarily large lower bounds in Theorem \ref{thm::asym_srisk}, as illustrated
by the following example. Since $A\pinv = \Tr(A\tran A)\inv \cdot A\tran$ for a
rank one matrix, \begin{align*}
  \left(\begin{bmatrix}
    p_1 & 0\\
    0 & p_2
  \end{bmatrix} \begin{bmatrix}
    X_{11} & 0\\
    X_{21} & 0
  \end{bmatrix}\right)\pinv \begin{bmatrix}
    p_1 & 0\\
    0 & p_2
  \end{bmatrix} = \begin{bmatrix}
    p_1 \cdot X_{11} & 0\\
    p_2 \cdot X_{21} & 0
  \end{bmatrix}\pinv \begin{bmatrix}
    p_1 & 0\\
    0 & p_2
  \end{bmatrix} = \begin{bmatrix}
    \tfrac{p_1^2 \cdot X_{11}}{(p_1 \cdot X_{11})^2 + (p_2 \cdot X_{21})^2} &
    \tfrac{p_2^2 \cdot X_{21}}{(p_1 \cdot X_{11})^2 + (p_2 \cdot X_{21})^2}\\
    0 & 0
  \end{bmatrix}.
\end{align*} Whenever $\abs[\big]{X_{11}} \ll \abs[\big]{X_{21}} \approx 1$, the
second data point is more informative regarding the value of $\starweight$. If
$\Sigma_{\bfeps} = I_2$, randomly weighted SGD with $\bbP\big(D_{11} = 1\big) =
p_1 \approx 1$ and $\bbP\big(D_{22} = 1\big) = p_2 \approx 0$ now causes the
risk of $\whweight$ to blow up in comparison with $X\pinv \bfY$ since
\begin{align*}
  \whX\pinv \sqrt{M_2} = \begin{bmatrix}
    \tfrac{p_1^2 \cdot X_{11}}{(p_1 \cdot X_{11})^2 + (p_2 \cdot X_{21})^2} &
    \tfrac{p_2^2 \cdot X_{21}}{(p_1 \cdot X_{11})^2 + (p_2 \cdot X_{21})^2}\\
    0 & 0
  \end{bmatrix} &\approx \begin{bmatrix}
    \tfrac{1}{X_{11}} & 0\\
    0 & 0
  \end{bmatrix}\\
  X\pinv = \begin{bmatrix}
    X_{11} & 0 \\
    X_{21} & 0
  \end{bmatrix}\pinv = \begin{bmatrix}
    \tfrac{X_{11}}{X_{11}^2 + X_{21}^2} & \tfrac{X_{21}}{X_{11}^2 + X_{21}^2} \\
    0 & 0
  \end{bmatrix} &\approx \begin{bmatrix}
    0 & \tfrac{1}{X_{21}} \\
    0 & 0
  \end{bmatrix}.
\end{align*} Despite the two data points being linearly dependent, meaning they
should in principle contain the same geometric information about $\starweight$,
prioritizing training on the ``wrong'' data point leads to bad statistical
performance of the iterates $\whweight_k$.

A weighting scheme that favors rows of $X$ with larger magnitudes, as introduced
in the beginning of this section, naturally avoids the pathology of the previous
example. However, the statistical noise $\bfeps$ may still result in better
recovery of $\starweight$ despite a ``bad'' weighting distribution. Suppose
again that $\Sigma_{\bfeps}$ is diagonal, but not equal to the identity matrix,
then the previous calculation in the rank-one setting leads to \begin{align*}
  \Tr\big(X\pinv \Sigma_{\bfeps} (X\pinv)\tran\big) &\approx
  \dfrac{\Sigma_{\bfeps, 22}}{X_{21}^2} \approx \Sigma_{\bfeps, 22}\\
  \Tr\Big(\big(\whX\pinv M_2^{1 / 2}\big) \Sigma_{\bfeps} \big(\whX\pinv
  M_2^{1 / 2}\big)\tran\Big) &\approx \dfrac{\Sigma_{\bfeps, 11}}{X_{11}^2}.
\end{align*} If $\Sigma_{\bfeps, 11}$ is sufficiently large in comparison with
$\Sigma_{\bfeps, 22}$, then iterates prioritizing the less informative data
point will outperform a uniform weighting distribution.

\begin{figure}[tb]
  \centering
  \includegraphics[width = 0.49\textwidth]{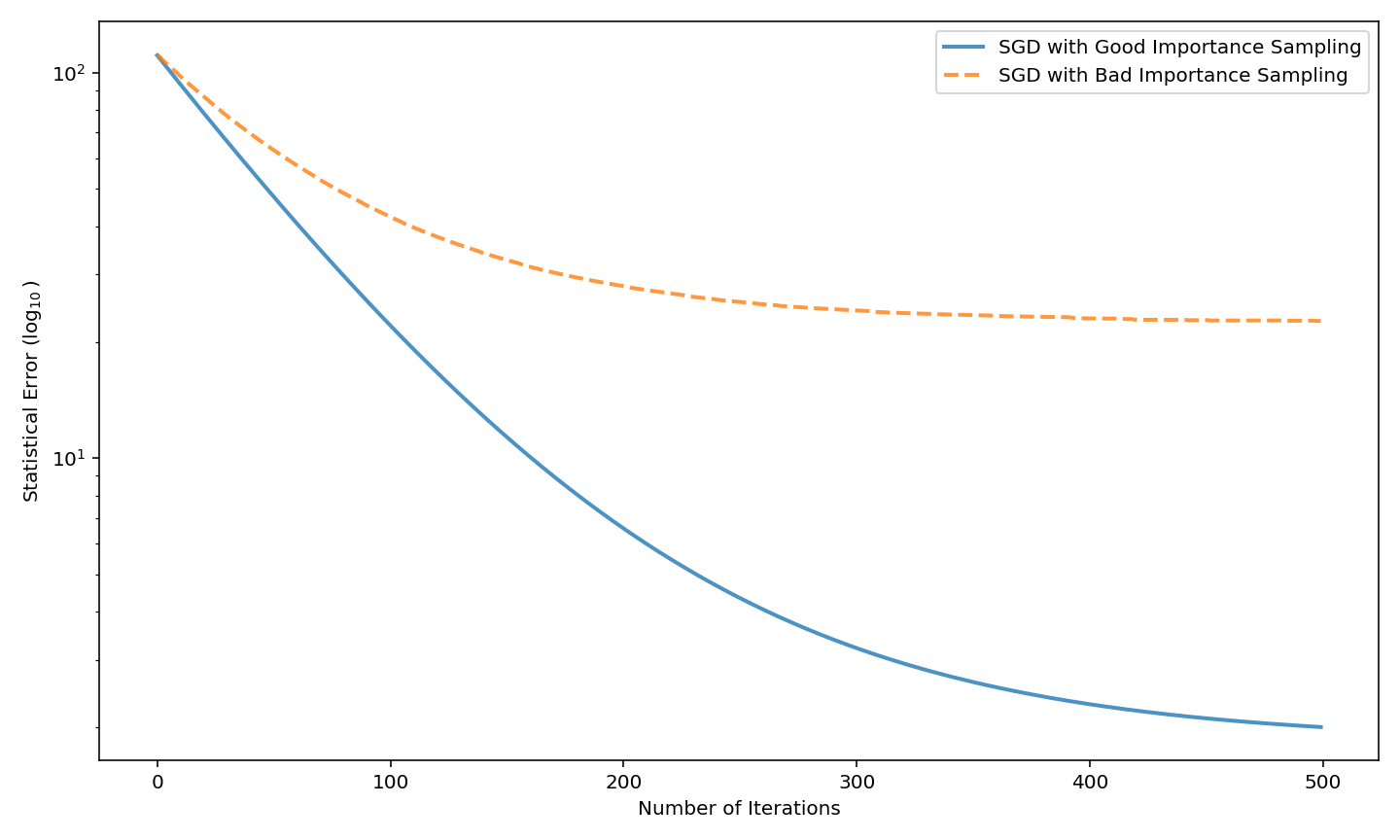}
  \includegraphics[width = 0.49\textwidth]{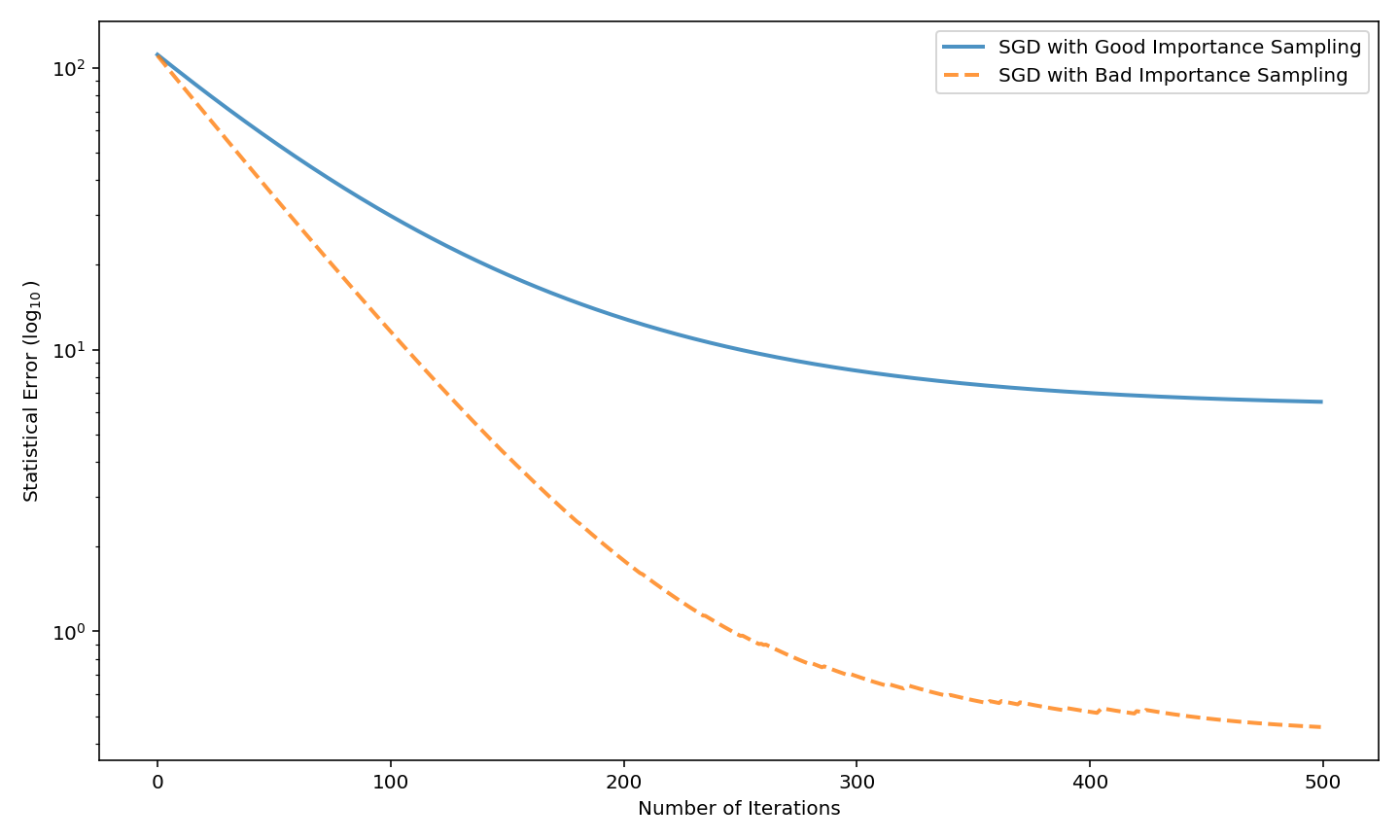}
  \caption{Comparison of statistical error $\E\big[\norm{\whweight_k -
  \starweight}_2^2\big]$ between ``good'' importance sampling $p_i =
  \exp\big(\norm{\bfX_i}_2\big) / \sum_{i = 1}^n \exp\big(\norm{\bfX_i}_2\big)$
  and ``bad'' importance sampling $p_i = \exp\big(- \norm{\bfX_i}_2\big) \cdot
  \sum_{i = 1}^n \exp\big(\norm{\bfX_i}_2\big)$. On the left-hand side, the data
  points with large norm feature small statistical noise $\eps_i$; the
  right-hand side features the same data matrix $X$, but now the important data
  points have large statistical noise.}
  \label{fig::stat_comparison}
\end{figure}

This illustrates a point of tension between the goals of fast optimization and
statistically optimal estimation. In common statistical usage, the weighted
least squares estimator $\whweight$ is employed to reduce the variance of
estimates based on data points of differing reliability. For uncorrelated
statistical noise $\bfeps$, this relies on weighting each data point according
to an estimate of $\Sigma_{\bfeps, ii}\inv$, see \cite{strutz_2016}. A weighting
scheme that prioritizes the wrong data points may accidentally amplify the
residual variances, as shown in the example above and in Figure
\ref{fig::stat_comparison} for the randomly weighted iterates. This underlines
how choices of optimization hyper-parameters, such as the sampling distribution
of data points, can effectively change the underlying statistical estimation
problem to cause undesirable outcomes.

\section{Discussion and Outlook}
\label{sec::disc}

We conclude with a discussion of some possible directions for future research.
One of the main limitations of our current study lies in the model choice.
Linear regression does not encapsulate all the complexities of optimizing the
empirical risk of a non-linear deep neural network. In general, a depth $L \geq
2$ network takes the form \begin{align*}
  f(\bfx) = \sigma_{W_L, \bfv_L} \circ \sigma_{W_{L - 1}, \bfv_{L - 1}} \circ
  \cdots \circ \sigma_{W_2, \bfv_2} \circ \sigma_{W_{1}, \bfv_{1}}(\bfx),
\end{align*} with $\sigma_{W_\ell, \bfv_\ell}(\bfx) = \sigma_\ell\big(W_\ell
\bfx + \bfv_\ell\big)$ and $\sigma_\ell$, $\ell = 1, \ldots, L$ a collection of
(usually non-linear) activation functions. Training such a network through ERM
requires estimating the entries of each affine transformation $\bfx \mapsto
W_\ell \bfx + \bfv_\ell$, which defines a difficult non-convex optimization
problem \cite{li_xu_et_al_2018}. As an intermediate step in terms of model
complexity, the study of matrix factorizations $\bfx \mapsto W_L W_{L - 1}
\cdots W_2 W_1 \bfx$ has experienced recent progress
\cite{arora_cohen_et_al_2019a, achour_malgouyres_et_al_2024,
chatterji_long_et_al_2022}. In particular, full-batch gradient descent shows
implicit bias towards nuclear norm minimal solutions when learning two-layer
positive semi-definite factorizations \cite{gunasekar_woodworth_et_al_2017}.
Hence, one may conjecture that a weighted nuclear norm minimal estimator plays
the same role as the weighted linear least squares estimator $\whweight$ does in
our analysis. Let $X$ and $Y$ respectively denote the matrices containing the
observed data points and the vector-valued labels, then two-layer loss with
random weighting takes the form \begin{align*}
  \big(W_1, W_2\big) \mapsto \norm[\big]{Y D - W_2 W_1 X D}_F^2,
\end{align*} where $\norm{\ \cdot\ }_F$ denotes the norm of a matrix when
treated as a vector. This leads to the interlinked random dynamical systems
\begin{align*}
  W_1(k + 1) &= W_1(k) - \dfrac{\alpha_k}{2} \cdot \nabla_{W_1(k)} \norm[\big]{Y
  D_k - W_2(k) W_1(k) X D_k}_F^2\\
  &= W_1(k) + \alpha_k \cdot W_2\tran(k) \Big(Y - W_2(k) W_1(k) X\Big) D^2_k
  X\tran\\
  W_2(k + 1) &= W_2(k) - \dfrac{\alpha_k}{2} \cdot \nabla_{W_2(k)} \norm[\big]{Y
  D_k - W_2(k) W_1(k) X D_k}_F^2\\
  &= W_2(k) + \alpha_k \cdot \Big(Y - W_2(k) W_1(k) X\Big) D^2_k
  X\tran W_1\tran(k).
\end{align*} Taking the conditional expectation and defining $\whX = X M_2^{1 /
2}$ and $\widehat{Y} = Y M_2^{1 / 2}$, one may analyze the underlying gradient
descent trajectory of the expected loss via the same technique as in
\cite{nguegnang_rauhut_et_al_2024}. In contrast, convergence of the second
moment seems more complicated. Due to the factorized structure of the loss,
$W_1(k)$ appears squared in the gradient with respect to $W_2(k)$ and vice
versa. This causes the second moment of the parameters to evolve differently
than the approximately affine dynamical structure shown in Lemma
\ref{lem::var_op_iso}, see also the corresponding discussion in Section 5 of
\cite{clara_langer_et_al_2024}. Further, the stationary distribution of the
iterates may not be unique; different valleys in the non-convex loss landscape
can trap the iterates with probability depending on the initialization.
Consequently, any stationary distribution of the iterates only admits local
uniqueness inside the basin of attraction of such a valley.

As a non-linear model with potentially interesting behavior under random
weightings, we may consider the weight tied auto-encoder studied in
\cite{ghosh_frei_et_al_2025}. Consider the empirical risk \begin{align*}
  \bfw \mapsto \dfrac{1}{n} \cdot \sum_{i = 1}^n \norm[\big]{\bfX_i - \bfw \cdot
  \sigma\big(\bfw\tran \bfX_i\big)}_2^2,
\end{align*} with $\sigma(x) = \max \{0, x\}$ the rectified linear unit (ReLU)
activation. As shown in \cite{ghosh_frei_et_al_2025}, mini-batch SGD with a
constant step-sizes manages to asymptotically find a global minimum of this
loss, but the minimum reached depends on the batch size. Accordingly, one may
expect that a weighted version of this result holds for SGD with biased
sampling. 

In the present article, the random weightings are sampled i.i.d., but in
practice it may be desirable to introduce dependency across iterations. As an
example, one could recompute the weighting distribution after every iteration to
identify data points that are most important to update during the following
iteration, which is reminiscent of methods such as saliency guided training
\cite{ismail_feizi_et_al_2021} and has applications in adaptive feature
decorrelation \cite{fröhlich_durst_et_al_2025}. Suppose we sample the diagonal
of $D_k$ from a categorical distribution with weights $p_1(k), \ldots, p_n(k)$
computed as a function of the past $\whweight_1, \ldots, \whweight_k$. The
resulting iterates do not necessarily minimize an expected loss in the form
\eqref{eq::loss_exp} and we may expect that any stationary distribution reached
by the iterates strongly depends on the initialization and outcomes of the
random weightings during early iterations. Further, recall from the proof of
Lemma \ref{lem::conv_exp_iso} that the evolution of the second moments of the
iterates can be computed via the law of total covariance. Dependency across
iterations of the weighting distribution then demands repeated application of
said law, which leads to many more terms that must be computed. As a starting
point, one may consider the case where the weights $p_1(k), \ldots, p_n(k)$ only
depend on $\whweight_k$, so that the iterates still evolve as a Markov process.
The weighted data points $\whX$ and labels $\whY$ should then be replaced with
iteration dependent counterparts, where the weighting is given as a function of
the previous iteration. We leave the details to future work.

%%%%%%%%%%%%%%%%%%%%%%%%%%%%%%%%%%%%%%%%%%%%%%%%%%%%%%%%%%%%%%%%%%%%%%%%%%%%%%%%
%%%%%%%%%%%%%%%%%%%%%%%%%%%%%%%%%%%%%%%%%%%%%%%%%%%%%%%%%%%%%%%%%%%%%%%%%%%%%%%%

{\sloppy
\printbibliography}

%%%%%%%%%%%%%%%%%%%%%%%%%%%%%%%%%%%%%%%%%%%%%%%%%%%%%%%%%%%%%%%%%%%%%%%%%%%%%%%%
%%%%%%%%%%%%%%%%%%%%%%%%%%%%%%%%%%%%%%%%%%%%%%%%%%%%%%%%%%%%%%%%%%%%%%%%%%%%%%%%

\appendix

\section{Proofs for Section \ref{sec::grad_desc}}
\label{sec::grad_desc_proof}

\subsection{Proof of Lemma \ref{lem::conv_gd}}

As shown in \eqref{eq::rec_gd_nn}, $\bfw_{k + 1} - X\pinv \bfY = (I - \alpha_k
\cdot \bbX) (\bfw_k - X\pinv \bfY)$ for every $k$. Combining the assumption
$\bfw_1 \perp \ker(X)$ with Lemma \ref{lem::pinv_c} shows that $\bfw_1 - X\pinv
\bfY \perp \ker(X)$. Since $\sup_{\ell} \alpha_\ell \cdot \norm{\bbX} < 1$,
Lemma \ref{lem::fixp_conv} and induction on $k$ then yield $\bfw_k - X\pinv \bfY
\perp \ker(X)$ for all $k$ while giving the desired estimate \begin{align*}
  \norm[\big]{\weight_{k + 1} - X\pinv \bfY}_2 \leq \left(\prod_{\ell = 1}^k
  \big(1 - \alpha_\ell \cdot \sigminp(\bbX)\big)\right) \cdot
  \norm[\big]{\weight_1 - X\pinv \bfY}_2.
\end{align*} To complete the proof it now suffices to apply Lemma
\ref{lem::misc_a}.

\section{Proofs for Section \ref{sec::conv_res_iso}}

\subsection{Proof of Lemma \ref{lem::conv_exp_iso}}

Combining the conditional expectations \eqref{eq::exp_lin} and
\eqref{eq::exp_shift} with the VAR representation \eqref{eq::gd_var},
independence of the random weighting matrices $D_k$ implies \begin{align}
  \label{eq::exp_rec}
  \E_D\big[\whweight_{k + 1} - \whweight\big] = \big(I - \alpha_k \cdot
  \whbbX\big) \E_D\big[\whweight_k - \whweight\big]
\end{align} for every $k \geq 1$. From here, the proof follows the same steps as
the proof of Lemma \ref{lem::conv_gd}. Assumptions \ref{ass::conv_b} and
\ref{ass::conv_c}, as well as Lemma \ref{lem::pinv_b} show that
$\E_D\big[\whweight_1 - \whweight\big] \perp \ker(\whX)$, so we may use Lemma
\ref{lem::fixp_conv} and induction on $k$ to show that $\E_D\big[\whweight_k -
\whweight\big] \perp \ker(\whX)$ for all $k$. Lemma \ref{lem::fixp_conv} also
gives the estimate \begin{align*}
  \norm[\Big]{\E_D\big[\whweight_{k + 1} - \whweight\big]}_2 \leq
  \left(\prod_{\ell = 1}^k \Big(1 - \alpha_\ell \cdot
  \sigminp(\whbbX)\Big)\right) \cdot \norm[\big]{\whweight_1 - \whweight}_2
\end{align*} with each $1 - \alpha_\ell \cdot \sigminp(\whbbX)$ contained in
$(0, 1)$ due to Assumption \ref{ass::conv_a}. Together with Lemma
\ref{lem::misc_a}, this finishes the proof.

\subsection{Proof of Lemma \ref{lem::var_op_iso}}

Throughout this proof, we write $A_k$ for the second moment of $\whweight_k -
\whweight$ with respect to $\E_D$. For any random vectors $\bfU$ and $\bfV$, the
law of total covariance yields \begin{align}
  \E\big[\bfU \bfU\tran\big] &= \Cov(\bfU) + \E[\bfU] \E[\bfU]\tran \nonumber\\
  &= \E\big[\Cov(\bfU \mid \bfV)\big] + \Cov\big(\E[\bfU \mid \bfV]\big) +
  \E[\bfU] \E[\bfU]\tran \nonumber\\
  &= \E\big[\Cov(\bfU \mid \bfV)\big] + \E\big[\E[\bfU \mid \bfV] \E[\bfU \mid
  \bfV]\tran\big] \label{eq::total_cov_id}
\end{align} Taking $\bfU = \whweight_{k + 1} - \whweight$ and $\bfV =
\whweight_k$ and employing the same arguments that led to \eqref{eq::exp_lin}
and \eqref{eq::exp_shift} yields \begin{align*}
  \E_D\big[\whweight_{k + 1} - \whweight \mid \whweight_k\big] = \E_D\big[I -
  \alpha_k \cdot X\tran D_k^2 X\big] \big(\whweight_k - \whweight\big) +
  \alpha_k \cdot \E_D\big[X\tran D_k^2 \big(\bfY - X \whweight\big) \mid
  \whweight_k \big] = \big(I - \alpha_k \cdot \whbbX\big) \big(\whweight_k -
  \whweight\big).
\end{align*} Writing $\Cov_D$ for the covariance with respect to $\E_D$,
\eqref{eq::total_cov_id} may then be rewritten as \begin{align}
  \label{eq::var_rec_cond}
  A_{k + 1} = \E_D\Big[\Cov_D\big(\whweight_{k + 1} - \whweight \mid
  \whweight_k\big)\Big] + \big(I - \alpha_k \cdot \whbbX\big) A_k \big(I -
  \alpha_k \cdot \whbbX\big).
\end{align} By definition, both $\bfY$ and $\whweight$ are constant with respect
to the randomness induced via the weighting matrices $D_k \sim D$. Using the VAR
representation \eqref{eq::gd_var}, the conditional covariance then simplifies to
\begin{align}
  \Cov_D\big(\whweight_{k + 1} - \whweight \mid \whweight_k\big) &=
  \Cov_D\Big(\big(I - \alpha_k \cdot X\tran D_k^2 X\big) \big(\whweight_k -
  \whweight\big) + \alpha_k \cdot X\tran D_k^2 \big(\bfY - X \whweight\big)
  \bigmid \whweight_k\Big) \nonumber\\
  &= \alpha_k^2 \cdot X\tran \Cov_D\bigg( D_k^2 \Big( - X \big(\whweight_k -
  \whweight\big) + \bfY - X \whweight\Big) \bigmid \whweight_k\bigg) X.
  \label{eq::cov_cond}
\end{align} To compute the latter expression, we proceed by proving a short
technical lemma.

\begin{lemma}
  \label{lem::cov_diag}
  Let a deterministic vector $\bfu$ and a random diagonal matrix $D$ of matching
  dimension be given. Suppose $\E[D^p]$ exists for each $p = 1, \ldots, 4$ and
  write $\Sigma_D$ for the covariance matrix of the vector with entries
  $D_{ii}^2$, then \begin{align*}
    \Cov\big(D^2 \bfu\big) = \Sigma_D \odot \bfu \bfu\tran,
  \end{align*} where $A \odot B$ denotes the element-wise product $(A \odot
  B)_{ij} = A_{ij} B_{ij}$.
\end{lemma}

\begin{proof}
  Using the definition of the covariance matrix of a random vector,
  \begin{align*}
    \Cov\big(D^2 \bfu\big) = \E\big[D^2 \bfu \bfu\tran D^2\big] - \E\big[D^2
    \bfu\big] \E\big[D^2 \bfu\big]\tran.
  \end{align*} The entries of the right-hand side matrices satisfy
  \begin{align*}
    E\big[D^2 \bfu \bfu\tran D^2\big]_{ij} &= \begin{cases}
      \E\big[D^4_{ii}\big] \cdot \big(\bfu \bfu\tran\big)_{ii} & \mbox{ if } i =
      j\\
      \E\big[D^2_{ii} D^2_{jj}\big] \cdot \big(\bfu \bfu\tran\big)_{ij} & \mbox{
      if } i \neq j
    \end{cases}\\
    \Big(\E\big[D^2 \bfu\big] \E\big[D^2 \bfu\big]\tran\Big)_{ij} &=
    \begin{cases}
      \E\big[D^2_{ii}\big]^2 \cdot \big(\bfu \bfu\tran\big)_{ii} & \mbox{ if } i
      = j\\
      \E\big[D^2_{ii}\big] \E\big[D^2_{jj}\big] \cdot \big(\bfu
      \bfu\tran\big)_{ij} & \mbox{ if } i \neq j.
    \end{cases}
  \end{align*} Subtracting the respective entries now yields the claimed
  element-wise identity \begin{align*}
    \Cov\big(D^2 \bfu\big)_{ij} = \Big(\E\big[D^2_{ii} D^2_{jj}\big] -
    \E\big[D_{ii}^2\big] \E\big[D_{jj}^2\big]\Big) \cdot \big(\bfu
    \bfu\tran\big)_{ij} = \Cov\big(D_{ii}^2, D_{jj}^2\big) \cdot \big(\bfu
    \bfu\tran\big)_{ij},
  \end{align*} where the right-hand side equals the $(i, j)$-entry of $\Sigma_D
  \odot \bfu \bfu\tran$.
\end{proof}

We now apply this lemma with $\bfu = - X (\whweight_k - \whweight) + \bfY - X
\whweight$, which is independent of $D_k$ when conditioning on $\whweight_k$, so
\eqref{eq::cov_cond} evaluates to \begin{align*}
  \alpha_k^2 \cdot X\tran \Cov_D\big(\whweight_{k + 1} - \whweight \mid
  \whweight_k\big) X &= \alpha_k^2 \cdot X\tran \Bigg(\Sigma_D \odot \Big(X
  \big(\whweight_k - \whweight\big) \big(\whweight_k - \whweight\big)\tran
  X\tran + \big(\bfY - X \whweight\big) \big(\bfY - X
  \whweight\big)\tran\Big)\Bigg) X\\
  &\qquad - \alpha_k^2 \cdot X\tran \Bigg(\Sigma_D \odot \Big(X \big(\whweight_k
  - \whweight\big) \big(\bfY - X \whweight\big)\tran + \big(\bfY - X
  \whweight\big) \big(\whweight_k - \whweight\big)\tran X\tran\Big)\Bigg) X
\end{align*} with $\Sigma_D$ the covariance matrix of $\big(D_{11}^2, \ldots,
D_{dd}^2\big)$, as defined in Lemma \ref{lem::cov_diag}. Combining this
computation with \eqref{eq::var_rec_cond} and exchanging the expectation
$\bbE_D$ with the linear operator $A \mapsto X\tran \big(\Sigma_D \odot X A
X\tran \big) X$ now results in the recursion \begin{align}
  A_{k + 1} &= \big(I - \alpha_k \cdot \whbbX\big) A_k \big(I - \alpha_k \cdot
  \whbbX\big) + \alpha_k^2 \cdot X\tran \Bigg(\Sigma_D \odot \bigg(X A_k X\tran
  + \big(\bfY - X \whweight\big) \big(\bfY - X \whweight\big)\tran\bigg)\Bigg) X
  \nonumber\\
  &\qquad - \E_D\Bigg[\alpha_k^2 \cdot X\tran \Bigg(\Sigma_D \odot \Big(X
  \big(\whweight_k - \whweight\big) \big(\bfY - X \whweight\big)\tran +
  \big(\bfY - X \whweight\big) \big(\whweight_k - \whweight\big)\tran
  X\tran\Big)\Bigg) X\Bigg] \nonumber\\
  &= S_{\alpha_k}\big(A_k\big) - \underbrace{\alpha_k^2 \cdot X\tran
  \Bigg(\Sigma_D \odot \bigg(X \E_D\Big[\big(\whweight_k - \whweight\big)
  \big(\bfY - X \whweight\big)\tran\Big] + \E_D\Big[\big(\bfY - X \whweight\big)
  \big(\whweight_k - \whweight\big)\tran\Big] X\tran\Big)\Bigg) X}_{= - \rho_k}.
  \label{eq::rec_op_rem}
\end{align} Symmetry of $\rho_k$ and the inclusion $\ker(X) \subset
\ker(\rho_k)$ follow directly from the previous display, so it remains to
estimate the norm of this remainder term. To this end, sub-multiplicativity of
the spectral norm and Lemma \ref{lem::misc_c} yield \begin{align*}
  \norm[\big]{\rho_k} \leq 2 \alpha_k^2 \cdot \norm{X}^3 \cdot \norm{\Sigma_D}
  \cdot \norm[\Bigg]{\E_D\Big[\big(\whweight_k - \whweight\big) \big(\bfY - X
  \whweight\big)\tran\Big]}
\end{align*} To complete the proof, we may now apply Lemma \ref{lem::misc_b} and
Lemma \ref{lem::conv_exp_iso} to estimate \begin{align*}
  \norm[\Bigg]{\E_D\Big[\big(\whweight_k - \whweight\big) \big(\bfY - X
  \whweight\big)\tran\Big]} &\leq \norm[\Big]{\E_D\big[\whweight_k -
  \whweight\big]}_2 \cdot \norm[\big]{\bfY - X \whweight}_2\\
  &\leq \left(\prod_{\ell = 1}^{k - 1} \Big(1 - \alpha_\ell \cdot
  \sigminp(\whbbX)\Big)\right) \cdot \norm[\big]{\whweight_1 - \whweight}_2
  \cdot \norm[\big]{\bfY - X \whweight}_2
\end{align*} which under Assumption \ref{ass::conv_a} vanishes as $k \to \infty$
by Lemma \ref{lem::misc_a}.

\subsection{Proof of Theorem \ref{thm::var_conv_iso}}

As in the proof of Lemma \ref{lem::var_op_iso}, we write $A_k =
\E_D\big[(\whweight_k - \whweight) (\whweight_k - \whweight)\tran\big]$. The
sequence of second moments satisfies $A_{k + 1} = S_{\alpha_k}(A_k) + \rho_k$,
with the remainder term $\rho_k$ as computed in \eqref{eq::rec_op_rem} vanishing
as $k \to \infty$. Using induction on $k$, we start by proving that
\begin{align}
  \label{eq::rec_unfold}
  A_{k + 1} = \Slin_{\alpha_k} \circ \cdots \circ \Slin_{\alpha_1}\big(A_1\big)
  + \sum_{\ell = 1}^k \Slin_{\alpha_k} \circ \cdots \circ \Slin_{\alpha_{\ell +
  1}} \big(\Sint_{\alpha_\ell}\big) + \sum_{m = 1}^k \Slin_{\alpha_k} \circ
  \cdots \circ \Slin_{\alpha_{m + 1}}\big(\rho_m\big),
\end{align} with the convention that the empty composition $\Slin_{\alpha_k}
\circ \cdots \Slin_{\alpha_{k + 1}}$ gives the identity operator. By definition,
each $\Slin_{\alpha_k}$ is a linear operator for every $k$ and so $A_2 =
S_{\alpha_1}(A_1) + \rho_1$ equals $\Slin_{\alpha_1}(A_1) + \Sint_{\alpha_1} +
\rho_1$, proving the base case. Suppose the result holds up to some $k \geq 1$,
then Lemma \ref{lem::var_op_iso} and linearity of $\Slin_{\alpha_k}$ imply
\begin{align*}
  A_{k + 1} &= S_{\alpha_k}\big(A_k\big) + \rho_k\\
  &= \Slin_{\alpha_k}\left(\Slin_{\alpha_{k - 1}} \circ \cdots \circ
  \Slin_{\alpha_1}\big(A_1\big) + \sum_{\ell = 1}^{k - 1} \Slin_{\alpha_{k - 1}}
  \circ \cdots \circ \Slin_{\alpha_{\ell + 1}} \big(\Sint_{\alpha_\ell}\big) +
  \sum_{m = 1}^{k - 1} \Slin_{\alpha_{k - 1}} \circ \cdots \circ
  \Slin_{\alpha_{m + 1}}\big(\rho_m\big)\right) + \Sint_{\alpha_k} + \rho_k\\
  &= \Slin_{\alpha_k} \circ \cdots \circ \Slin_{\alpha_1}\big(A_1\big) +
  \sum_{\ell = 1}^{k - 1} \Slin_{\alpha_k} \circ \cdots \circ
  \Slin_{\alpha_{\ell + 1}} \big(\Sint_{\alpha_\ell}\big) + \sum_{m = 1}^{k - 1}
  \Slin_{\alpha_k} \circ \cdots \circ \Slin_{\alpha_{m + 1}}\big(\rho_m\big) +
  \Sint_k + \rho_k\\
  &= \Slin_{\alpha_k} \circ \cdots \circ \Slin_{\alpha_1}\big(A_1\big) +
  \sum_{\ell = 1}^k \Slin_{\alpha_k} \circ \cdots \circ \Slin_{\alpha_{\ell +
  1}} \big(\Sint_{\alpha_\ell}\big) + \sum_{m = 1}^k \Slin_{\alpha_k} \circ
  \cdots \circ \Slin_{\alpha_{m + 1}}\big(\rho_m\big).
\end{align*} This proves the induction step and so \eqref{eq::rec_unfold} holds
for all $k \geq 0$. To proceed, we require a bound on the effective operator
norms of the linear operators $\Slin_{\alpha_k}$ when applied to $A_0$ and
$\rho_m$.

\begin{lemma}
  \label{lem::lin_op_bound}
  Fix a symmetric $(d \times d)$-matrix $A$ with $\ker(X) \subset \ker(A)$ and
  \begin{align*}
    \alpha < \left\{\dfrac{1}{\norm{\whbbX}},
    \dfrac{\sigminp(\whbbX)}{\sigminp(\whbbX)^2 + \norm{X}^4 \cdot
    \norm{\Sigma_D}}\right\},
  \end{align*} then \begin{align*}
    \norm[\Big]{\Slin_{\alpha}(A)} \leq \big(1 - \alpha \cdot
    \sigminp(\whbbX)\big) \cdot \norm{A}.
  \end{align*}
\end{lemma}

\begin{proof}
  As $A$ is symmetric and $\Slin_\alpha$ maps the space of symmetric matrices to
  itself, the singular values of $\Slin_\alpha(A)$ are the magnitudes of its
  eigenvalues. To bound the latter, we will use the variational characterization
  of the eigenvalues, see Theorem 4.2.6 in \cite{horn_johnson_2013}. Fix a
  non-zero unit vector $\bfw$. The definition of $\Slin_\alpha$ entails
  \begin{align}
    \label{eq::slin_eig}
    \bfw\tran \Slin_\alpha(A) \bfw = \bfw\tran \big(I - \alpha \cdot \whbbX\big)
    A \big(I - \alpha \cdot \whbbX\big) \bfw + \alpha^2 \cdot \bfw\tran X\tran
    \Big(\Sigma_D \odot \big(X A X\tran\big)\Big) X \bfw
  \end{align} If $\bfw \in \ker(X)$, then also $\bfw \in \ker(\whbbX)$, so $I -
  \alpha \cdot \whbbX$ acts as the identity on $\ker(X)$. Together with the
  assumption $\ker(X) \subset \ker(A)$, this reduces \eqref{eq::slin_eig} to
  \begin{align}
    \label{eq::slin_ker}
    \bfw\tran \Slin_\alpha(A) \bfw = \bfw\tran A \bfw = 0,
  \end{align} so all non-zero eigenvalues of $\Slin_\alpha(A)$ must correspond
  to eigenvectors in the orthogonal complement of $\ker(X)$. As $A$ is
  symmetric, it admits a singular value decomposition of the form $U \Sigma
  U\tran$, with $\Sigma$ a positive semi-definite $(d \times d)$-diagonal
  matrix. Accordingly, for any matrix $B$ the Cauchy-Schwarz inequality implies
  \begin{align}
    \label{eq::cent_mat_bd}
    \abs[\Big]{\bfw B\tran A B\tran \bfw} = \abs[\Big]{\bfw B\tran U
    \sqrt{\Sigma} U\tran U \sqrt{\Sigma} U\tran B\tran \bfw} \leq \norm[\Big]{U
    \sqrt{\Sigma} U\tran B \bfw}_2^2 \leq \norm[\big]{U \sqrt{\Sigma} U\tran}^2
    \cdot \norm[\big]{B \bfw}_2^2 = \norm{A} \cdot \norm[\big]{B \bfw}_2^2
  \end{align} Applying a similar argument to the singular value decomposition
  of the symmetric matrix $\Sigma_D \odot \big(X A X\tran\big)$, we also find
  that \begin{align*}
    \abs[\bigg]{\bfw\tran X\tran \Big(\Sigma_D \odot \big(X A X\tran\big)\Big) X
    \bfw} \leq \norm[\Big]{\Sigma_D \odot \big(X A X\tran\big)} \cdot
    \norm[\big]{X \bfw}_2^2.
  \end{align*} Suppose now that $\bfw \perp \ker(X)$, which implies $\bfw \perp
  \ker(\whbbX)$ by Assumption \ref{ass::conv_c}. Taking the absolute value in
  \eqref{eq::slin_eig}, inserting the previous computations, as well as applying
  Lemma \ref{lem::pinv_c} and Lemma \ref{lem::misc_c}, we arrive at the estimate
  \begin{align*}
    \abs[\Big]{\bfw\tran \Slin_\alpha(A) \bfw} &\leq \norm{A} \cdot
    \norm[\Big]{\big(I - \alpha \cdot \whbbX\big) \bfw}_2^2 + \alpha^2 \cdot
    \norm[\Big]{\Sigma_D \odot \big(X A X\tran\big)} \cdot \norm[\big]{X
    \bfw}_2^2\\
    &\leq \bigg(\big(1 - \alpha \cdot \sigminp(\whbbX)\big)^2 + \alpha^2 \cdot
    \norm{X}^4 \cdot \norm{\Sigma_D}\bigg) \cdot \norm{A},
  \end{align*} where we recall that the norm of $\bfw$ evaluates to $1$. It now
  suffices to further bound the scalar multiplying $\norm{A}$ in the previous
  display, then the variational characterization of eigenvalues completes the
  proof. Expanding the square and using the assumption on $\alpha$ yields the
  desired inequality \begin{align*}
    \big(1 - \alpha \cdot \sigminp(\whbbX)\big)^2 + \alpha^2 \cdot \norm{X}^4
    \cdot \norm{\Sigma_D} &= 1 - 2 \alpha \cdot \sigminp(\whbbX) + \alpha^2 \cdot
    \Big(\sigminp(\whbbX)^2 + \norm{X}^4 \cdot \norm{\Sigma_D}\Big)\\
    &\leq 1 - \alpha \cdot \sigminp(\whbbX).
  \end{align*}
\end{proof}

We now return to the expression \eqref{eq::rec_unfold} for $A_{k + 1}$, where we
will use Lemma \ref{lem::lin_op_bound} to bound the norm of each constituent
summand, which then yields an expression for $A_{k + 1}$ up to a vanishing
remainder.

\begin{lemma}
  \label{lem::var_ass}
  In addition to Assumption \ref{ass::conv_iso}, suppose $\sup_{\ell}
  \alpha_\ell$ satisfies the requirements of Lemma \ref{lem::lin_op_bound}.
  Then, for every $k \geq 1$ \begin{align*}
    \norm[\Bigg]{A_{k + 1} - \sum_{\ell = 1}^k \Slin_{\alpha_k} \circ \cdots
    \circ \Slin_{\alpha_{\ell + 1}} \big(\Sint_{\alpha_\ell}\big)} \leq
    C_0 \cdot \left(1 + \sum_{\ell = 1}^k \alpha_\ell^2\right)
    \cdot \left(\max_{m = 1 \ldots, k} \prod_{\substack{\ell = 1\\ \ell \neq
    m}}^{k} \Big(1 - \alpha_\ell \cdot \sigminp(\whbbX)\Big)\right),
  \end{align*} with constant $C_0 = \norm[\big]{(\whweight_1 - \whweight)
  (\whweight_1 - \whweight)\tran} + 2 \cdot \norm{X}^3 \cdot \norm{\Sigma_D}
  \cdot \norm[\big]{\whweight_1 - \whweight}_2 \cdot \norm[\big]{\bfY - X
  \whweight}_2$.
\end{lemma}

\begin{proof}
  As shown in \eqref{eq::slin_ker}, each linear operator $\Slin_{\alpha_\ell}$
  maps the space of matrices $A$ satisfying $\ker(X) \subset \ker(A)$ to itself.
  Due to Assumption \ref{ass::conv_b} and Lemma \ref{lem::var_op_iso}, this
  includes $A_1$ as well as each remainder term $\rho_m$. In turn, repeated
  application of Lemma \ref{lem::lin_op_bound} yields the estimates
  \begin{align*}
    \norm[\Big]{\Slin_{\alpha_k} \circ \cdots \circ \Slin_{\alpha_1}
    \big(A_1\big)} &\leq \left(\prod_{\ell = 1}^k \big(1 - \alpha_\ell \cdot
    \sigminp(\whbbX)\big)\right) \cdot \norm{A_1}\\
    \norm[\Big]{\Slin_{\alpha_k} \circ \cdots \circ \Slin_{\alpha_{m +
    1}}\big(\rho_m\big)} &\leq \left(\prod_{\ell = m + 1}^k \big(1 - \alpha_\ell
    \cdot \sigminp(\whbbX)\big)\right) \cdot \norm{\rho_m},
  \end{align*} valid for every $k \geq 1$ and $m < k$. Together with the
  estimate for $\norm{\rho_m}$ computed in Lemma \ref{lem::var_op_iso}, we may
  now rearrange \eqref{eq::rec_unfold} and take the norm to find the desired
  bound \begin{align*}
    &\norm[\Bigg]{A_{k + 1} - \sum_{\ell = 1}^k \Slin_{\alpha_k} \circ \cdots
    \circ \Slin_{\alpha_{\ell + 1}} \big(\Sint_{\alpha_\ell}\big)}\\
    \leq\ &\left(\prod_{\ell = 1}^k \big(1 - \alpha_\ell \cdot
    \sigminp(\whbbX)\big)\right) \cdot \norm{A_1} + \sum_{m = 1}^{k}
    \left(\prod_{\ell = m + 1}^k \big(1 - \alpha_\ell \cdot
    \sigminp(\whbbX)\big)\right) \cdot \norm{\rho_m}\\
    \leq\ &\Big(\norm{A_1} + 2 \cdot \norm{X}^3 \cdot \norm{\Sigma_D} \cdot
    \norm[\big]{\whweight_1 - \whweight}_2 \cdot \norm[\big]{\bfY - X
    \whweight}_2\Big) \cdot \left(1 + \sum_{\ell = 1}^k \alpha_\ell^2\right)
    \cdot \left(\max_{m = 1 \ldots, k} \prod_{\substack{\ell = 1\\ \ell \neq
    m}}^{k} \Big(1 - \alpha_\ell \cdot \sigminp(\whbbX)\Big)\right).
  \end{align*}
\end{proof}

We will first prove the statement regarding square summable step-sizes in
Theorem \ref{thm::var_conv_iso_a}. To this end, recall from Lemma
\ref{lem::var_op_iso} that \begin{align*}
  \Sint_{\alpha_\ell} = \alpha_\ell^2 \cdot X\tran \Bigg(\Sigma_D \odot
  \big(\bfY - X \whweight\big) \big(\bfY - X
  \whweight\big)\tran \Bigg) X.
\end{align*} The kernel of the constant symmetric matrix multiplying
$\alpha_\ell^2$ contains $\ker(X)$, so the triangle inequality and repeated
application of Lemma \ref{lem::lin_op_bound} result in \begin{equation}
  \label{eq::res_bound}
  \begin{split}
    \norm*{\sum_{\ell = 1}^k \Slin_{\alpha_k} \circ \cdots \circ
    \Slin_{\alpha_{\ell + 1}} \big(\Sint_{\alpha_\ell}\big)} &\leq \sum_{\ell =
    1}^k \left(\prod_{m = \ell + 1}^k \big(1 - \alpha_m \cdot
    \sigminp(\whbbX)\big)\right) \cdot \norm[\big]{\Sint_{\alpha_\ell}}\\
    &\leq \norm{X}^2 \cdot \norm{\Sigma_D} \cdot \norm[\big]{\bfY - X
    \whweight}_2^2 \cdot \sum_{\ell = 1}^k \alpha_\ell^2 \cdot \left(\prod_{m =
    \ell + 1}^k \big(1 - \alpha_m \cdot \sigminp(\whbbX)\big)\right)
  \end{split}
\end{equation} where the second inequality follows from sub-multiplicativity of
the norm and Lemma \ref{lem::misc_b} and \ref{lem::misc_c}. Defining $c_\ell =
\alpha_\ell \cdot \sigminp(\whbbX)$, the latter expression is proportional to
$\sum_{\ell = 1}^k c_\ell^2 \cdot \prod_{m = \ell + 1}^k (1 - c_m)$. By
construction, $c_\ell \in (0, 1)$ for every $\ell$, so Lemma \ref{lem::misc_a}
implies \begin{align*}
  \sum_{\ell = 1}^k c_\ell^2 \cdot \prod_{m = \ell + 1}^k (1 - c_m) &\leq
  \sum_{\ell = 1}^k c_{\ell}^2 \cdot \exp\left(- \sum_{\ell = m + 1}^k
  c_\ell\right)\\
  &\leq \sum_{\ell = \floor{k / 2}}^k c_\ell^2 + \exp\left(- \sum_{m =
  \ceil{k / 2}}^{k} c_m\right).
\end{align*} As $k \to \infty$, the tail series $\sum_{\ell = \floor{k / 2}}^{k}
c_\ell^2$ must vanish due to $c_{\ell}$ being square summable. The exponential
term must then also go to $0$ since the $c_\ell$ are non-summable. Accordingly,
\eqref{eq::res_bound} converges to $0$, which proves the statement about
square-summable step-sizes.

Suppose now that $\alpha_\ell = \alpha / \ell$ for some constant $\alpha > 0$,
then $\sum_{\ell = 1}^k \alpha_\ell > \alpha \cdot \log(k)$ for every $k \geq
1$. In particular, $\sum_{\ell = 1}^k \ell\inv - \log(k)$ converges to the
Euler-Mascheroni constant $\gamma = 0.577 \ldots$ from below. Consequently,
Lemma \ref{lem::misc_a} implies \begin{align*}
  \sum_{\ell = 1}^k \alpha_\ell^2 \cdot \left(\prod_{m = \ell + 1}^k \big(1 -
  \alpha_m \cdot \sigminp(\whbbX)\big)\right) &\leq \sum_{\ell = 1}^k
  \alpha_\ell^2 \cdot \exp \left(- \sigminp(\whbbX) \cdot \sum_{m = \ell + 1}^k
  \alpha_m\right)\\
  &= \exp \left(- \sigminp(\whbbX) \cdot \sum_{m = 1}^k \alpha_m\right) \cdot
  \sum_{\ell = 1}^k \alpha_\ell^2 \cdot \exp \left(\sigminp(\whbbX) \cdot
  \sum_{m = 1}^\ell \alpha_m\right)\\
  &\leq \exp\Big(- \alpha \sigminp(\whbbX) \cdot \log(k)\Big)
  \cdot \sum_{\ell = 1}^{k} \alpha_\ell^2 \cdot \exp \left(\sigminp(\whbbX)
  \cdot \sum_{m = 1}^{\ell} \alpha_m\right)\\
  &\leq \exp\Big(\alpha \sigminp(\whbbX) \cdot \big(\gamma - \log(k)\big)\Big)
  \cdot \sum_{\ell = 1}^k \alpha_\ell^2 \cdot \exp\Big(\alpha \sigminp(\whbbX)
  \cdot \log(\ell)\Big)\\
  &= e^{\alpha \sigminp(\whbbX) \gamma} \cdot \dfrac{1}{k^{\alpha
  \sigminp(\whbbX)}} \cdot \sum_{\ell = 1}^k \dfrac{\alpha}{\ell^{2 - \alpha
  \sigminp(\whbbX)}}
\end{align*} where the last inequality follows from. Recall that $\sum_{\ell =
1}^{k} 1 / \ell^{1 + a}$ converges to a finite constant for all $a > 0$, namely
$\zeta(1 + a)$, with the latter denoting the Riemann $\zeta$-function. Together
with \eqref{eq::res_bound}, we now arrive at the estimate \begin{align*}
  \norm*{\sum_{\ell = 1}^k \Slin_{\alpha_k} \circ \cdots \circ
  \Slin_{\alpha_{\ell + 1}} \big(\Sint_{\alpha_\ell}\big)} \leq \norm{X}^2 \cdot
  \norm{\Sigma_D} \cdot \norm[\big]{\bfY - X \whweight}_2^2 \cdot e^{\alpha
  \sigminp(\whbbX) \gamma} \cdot \alpha \zeta\big(2 - \alpha \cdot
  \sigminp(\whbbX)\big) \cdot \dfrac{1}{k^{\alpha \sigminp(\whbbX)}}
\end{align*} Combining the latter with Lemma \ref{lem::var_ass} and applying
Lemma \ref{lem::misc_a} now leads to the desired convergence rate \begin{align*}
  \norm[\big]{A_{k + 1}} \leq \norm[\Bigg]{A_{k + 1} - \sum_{\ell = 1}^k
  \Slin_{\alpha_k} \circ \cdots \circ \Slin_{\alpha_{\ell + 1}}
  \big(\Sint_{\alpha_\ell}\big)} + \norm*{\sum_{\ell = 1}^k \Slin_{\alpha_k}}
  \leq C_1 \cdot \dfrac{1}{k^{\alpha \sigminp(\whbbX)}}
\end{align*} with constant \begin{align*}
  C_1 = C_0 \cdot \left(1 + \alpha \cdot \dfrac{\pi^2}{6}\right)+ \norm{X}^2
  \cdot \norm{\Sigma_D} \cdot \norm[\big]{\bfY - X \whweight}_2^2 \cdot
  e^{\alpha \sigminp(\whbbX) \gamma} \cdot \alpha \zeta\big(2 - \alpha \cdot
  \sigminp(\whbbX)\big).
\end{align*} This concludes the proof of Theorem \ref{thm::var_conv_iso_a}.

We now turn our attention to the setting of constant step-sizes $\alpha_k =
\alpha$, as in Theorem \ref{thm::var_conv_iso_b}. In this case, the affine map
$S_{\alpha}$ in Lemma \ref{lem::var_op_iso} is the same for each iteration and
so Lemma \ref{lem::var_ass} implies \begin{align*}
  \norm[\Bigg]{A_{k + 1} - \sum_{\ell = 1}^k \big(\Slin_\alpha\big)^{\ell - 1}
  \big(\Sint_\alpha\big)} \leq C_0 \cdot \big(1 + k \alpha^2\big) \cdot \big(1
  - \alpha_\ell \cdot \sigminp(\whbbX)\big)^{k - 1}.
\end{align*} Recall from Lemma \ref{lem::var_op_iso} that $\Sint_\alpha$ is a
symmetric matrix, the kernel of which contains $\ker(X)$. The collection of
matrices with these properties is stable with respect to the usual vector space
operations and hence forms a complete metric space under the spectral norm. As
shown in Lemma \ref{lem::lin_op_bound}, the effective operator norm of
$\Slin_\alpha$ on this space is given by $1 - \alpha \cdot \sigminp(\whbbX) <
1$. Consequently, the restriction of $\id - \Slin_\alpha$ to this sub-space may
be inverted via its Neumann series (Proposition 5.3.4 in \cite{helemskii_2006},
which yields \begin{align*}
  \norm*{\sum_{\ell = 1}^k \big(\Slin_\alpha\big)^{\ell - 1}
  \big(\Sint_\alpha\big) - \big(\id - \Slin_\alpha\big)\inv_{\ker(X)}
  \big(\Sint_\alpha\big)} \leq \big(1 - \alpha \cdot \sigminp(\whbbX)\big)^k
  \norm[\Big]{\big(\id - \Slin_\alpha\big)\inv_{\ker(X)} \big(\Sint_\alpha\big)}
\end{align*} Combining these computations now leads to the desired convergence
rate \begin{align*}
  &\norm[\Bigg]{A_{k + 1} - \big(\id - \Slin_\alpha\big)\inv_{\ker(X)}
  \big(\Sint_\alpha\big)}\\
  \leq\ &\norm[\Bigg]{A_{k + 1} - \sum_{\ell = 1}^k \big(\Slin_\alpha\big)^{\ell
  - 1} \big(\Sint_\alpha\big)} + \norm*{\sum_{\ell = 1}^k
  \big(\Slin_\alpha\big)^{\ell - 1} \big(\Sint_\alpha\big) - \big(\id -
  \Slin_\alpha\big)\inv_{\ker(X)} \big(\Sint_\alpha\big)}\\
  \leq\ &C_2 \cdot \big(2 + k \alpha^2\big) \cdot \big(1 - \alpha \cdot
  \sigminp(\whbbX)\big)^{k - 1}
\end{align*} with constant \begin{align*}
  C_2 = C_0 + \norm[\Big]{\big(\id - \Slin_\alpha\big)\inv_{\ker(X)}
  \big(\Sint_\alpha\big)}.
\end{align*} Applying Lemma \ref{lem::misc_a} completes the proof of Theorem
\ref{thm::var_conv_iso_b}.

\subsection{Proof of Theorem \ref{thm::stat_dist}}

The proof of the first statement follows along similar lines as the proof of
Lemma 1 in \cite{li_schmidt-hieber_et_al_2024}. Due to Assumption
\ref{ass::gmc_b}, both $\norm{I - \alpha \cdot X\tran D^2 X}^q$ and
$\norm{X\tran D^2 (\bfY - X \whweight)}_2^q$ admit deterministic upper-bounds
that are polynomial in $\norm{\bbX}$, $\norm{\bfY}_2$, and $\tau^2$, with degree
depending on $q$. Consequently, $\norm{\whweight_k}_2^q$ has finite expectation
with respect to $D$ for all $k, q \geq 1$.

Fix $k \geq 1$, then the specific form of the gradient descent recursion
\eqref{eq::gd_var} and the coupling between $\bfu_{k + 1}$ and $\bfv_{k + 1}$
imply \begin{align}
  &\norm[\big]{\whu_{k + 1} - \whv_{k + 1}}_2^q \nonumber\\
  =\ &\norm[\big]{\whu_{k + 1} - \whweight + \whweight - \whv_{k + 1}}_2^q
  \nonumber\\
  =\ &\norm[\Bigg]{\big(I - \alpha \cdot X\tran D_k^2 X\big) \big(\whu_k -
  \whweight\big) + \alpha \cdot X\tran D_k^2 \big(\bfY - X \whweight\big) -
  \Bigg(\big(I - \alpha \cdot X\tran D_k^2 X\big) \big(\whv_k - \whweight\big) +
  \alpha \cdot X\tran D_k^2 \big(\bfY - X \whweight\big)\Bigg)}_2^q \nonumber\\
  =\ &\norm[\Big]{\big(I - \alpha \cdot X\tran D_k^2 X\big) \big(\whu_k -
  \whv_k\big)}_2^q. \label{eq::rec_diff_gmc}
\end{align} Due to Assumption \ref{ass::conv_a}, the initial difference $\whu_1
- \whv_1$ is almost surely orthogonal to $\ker(X)$. For any $\bfw \in \ker(X)$
and $k \geq 1$, \begin{align*}
  \bfw\tran \big(I - \alpha \cdot X\tran D_k^2 X\big) \big(\whu_k - \whv_k\big)
  = \bfw\tran \big(\whu_k - \whv_k\big) = \bfzero
\end{align*} so induction on $k$ proves that $\big(\whu_k - \whv_k\big) \perp
\ker(X)$ almost surely for all $k$. Taking the expectation with respect to $D$
in \eqref{eq::rec_diff_gmc}, we may divide and multiply the latter by
$\norm{\whu_k - \whv_k}^q_2$ to rewrite \begin{align*}
  &\E_D\Bigg[\norm[\Big]{\big(I - \alpha \cdot X\tran D_k^2 X\big) \big(\whu_k
  - \whv_k\big)}_2^q\Bigg]\\
  =\ &\E_D\Bigg[\tfrac{1}{\norm{\whu_k - \whv_k}^q_2} \cdot \norm[\Big]{\big(I -
  \alpha \cdot X\tran D_k^2 X\big) \big(\whu_k - \whv_k\big)}_2^q \cdot
  \norm[\big]{\whu_k - \whv_k}^q_2\Bigg]\\
  =\ &\E_D\left[\E_D\Bigg[\tfrac{1}{\norm{\whu_k - \whv_k}^q_2} \cdot
  \norm[\Big]{\big(I - \alpha \cdot X\tran D_k^2 X\big) \big(\whu_k -
  \whv_k\big)}_2^q \Bigmid \whu_k, \whv_k \Bigg] \cdot
  \norm[\big]{\whu_k - \whv_k}^q_2 \right].
\end{align*} Write $\widehat{\bfz}_k$ for the unit vector in the direction of
$\whu_k - \whv_k$, then the conditional expectation of $\norm{(I - \alpha \cdot
X\tran D_k^2 X) \widehat{\bfz}_k}_2^q$ is a deterministic function of
$\widehat{\bfz}_k$. To maximize it, we may take the supremum over the orthogonal
complement of $\ker(X)$. Since $D_k$ is generated independent of $\whu_k$ and
$\whv_k$, this results in \begin{align*}
  &\E_D\left[\E_D\Bigg[\norm[\Big]{\big(I - \alpha \cdot X\tran D_k^2 X\big)
  \widehat{\bfz}_k}_2^q \Bigmid \whu_k, \whv_k \Bigg] \cdot \norm[\big]{\whu_k -
  \whv_k}^q_2 \right]\\
  \leq\ &\E_D\left[\sup_{\substack{\norm{\bfw} = 1\\ \bfw \perp \ker(X)}}
  \E_D\Bigg[\norm[\Big]{\big(I - \alpha \cdot X\tran D_k^2 X\big) \bfw}_2^q
  \Bigmid \whu_k, \whv_k \Bigg] \cdot \norm[\big]{\whu_k - \whv_k}^q_2 \right]\\
  =\ &\sup_{\substack{\norm{\bfw} = 1\\ \bfw \perp \ker(X)}}
  \E_D\Bigg[\norm[\Big]{\big(I - \alpha \cdot X\tran D_k^2 X\big)
  \bfw}_2^q\Bigg] \cdot \E_D\Big[\norm[\big]{\whu_k - \whv_k}^q_2\Big].
\end{align*}

To prove the first claim, it now suffices to show that the supremum over $\bfw$
in the previous display always leads to a multiplier contained in $(0, 1)$. By
Assumption \ref{ass::gmc_c}, the singular values of $X\tran D_k^2 X$ lie in $[0,
1)$, meaning $\norm[\big]{I - \alpha \cdot X\tran D_k^2 X} \leq 1$. We first
treat the case $q \geq 2$, which allows for the upper bound \begin{align*}
  \E_D\Bigg[\norm[\Big]{\big(I - \alpha \cdot X\tran D_k^2 X\big)
  \bfw}_2^q\Bigg] &= \E_D\Bigg[\norm[\Big]{\big(I - \alpha \cdot X\tran D_k^2
  X\big) \bfw}_2^2 \cdot \norm[\Big]{\big(I - \alpha \cdot X\tran D_k^2 X\big)
  \bfw}_2^{q - 2}\Bigg]\\
  &\leq \E_D\Bigg[\norm[\Big]{\big(I - \alpha \cdot X\tran D_k^2 X\big)
  \bfw}_2^2 \cdot \norm[\big]{I - \alpha \cdot X\tran D_k^2 X}^{q - 2}\Bigg]\\
  &\leq \E_D\Bigg[\norm[\Big]{\big(I - \alpha \cdot X\tran D_k^2 X\big)
  \bfw}_2^2\Bigg].
\end{align*} Having reduced the problem to bounding the expectation of the
squared norm, we further note via symmetry of $X\tran D_k^2 X$ that
\begin{align*}
  \sup_{\substack{\norm{\bfw} = 1\\ \bfw \perp \ker(X)}}
  \E_D\Bigg[\norm[\Big]{\big(I - \alpha \cdot X\tran D_k^2 X\big)
  \bfw}_2^2\Bigg] &= \sup_{\substack{\norm{\bfw} = 1\\ \bfw \perp \ker(X)}}
  \bfw\tran \E_D\bigg[\big(I - \alpha \cdot X\tran D_k^2 X\big)^2\bigg] \bfw.
\end{align*} Assumption \ref{ass::gmc_b} and sub-multiplicativity of the norm
imply $\norm{D_k X X\tran D_k} \leq \tau^2 \cdot \norm{\bbX}$, so an argument
similar to \eqref{eq::cent_mat_bd} yields \begin{align*}
  \bfw\tran \bigg(- 2 \alpha \cdot \E_D\big[X\tran D_k^2 X\big] + \alpha^2 \cdot
  \E_D\Big[X\tran D_k^2 X X\tran D_k^2 X\Big]\bigg) \bfw \leq - \alpha \cdot
  \big(2 - \alpha \tau^2 \cdot \norm{\bbX}\big) \cdot \bfw\tran \whbbX \bfw
\end{align*} Expanding the square of $I - \alpha \cdot X\tran D_k^2 X$ and
combining Assumptions \ref{ass::gmc_b} and \ref{ass::gmc_c} with Lemma
\ref{lem::fixp_conv} now leads to the final bound \begin{align}
   \sup_{\substack{\norm{\bfw} = 1\\ \bfw \perp \ker(X)}} \bfw\tran
   \E_D\bigg[\big(I - \alpha \cdot X\tran D_k^2 X\big)^2\bigg] \bfw &\leq
   \sup_{\substack{\norm{\bfw} = 1\\ \bfw \perp \ker(X)}} \bfw\tran \bigg(I -
   \alpha \cdot \big(2 - \alpha \tau^2 \cdot \norm{\bbX}\big) \cdot \whbbX\bigg)
   \bfw \nonumber\\
   &= 1 - \alpha \cdot \big(2 - \alpha \tau^2 \cdot \norm{\bbX}\big) \cdot
   \sigminp(\whbbX). \label{eq::gmc_fact}
\end{align} Assumption \ref{ass::gmc_c} further implies $2 - \alpha \tau^2 \cdot
\norm{\bbX} < 2$. The singular values of $\whbbX$ are always bounded by $\tau^2
\cdot \norm{\bbX}$, so \eqref{eq::gmc_fact} lies strictly below $1$. The case $q
\in (1, 2)$ now follows from Hölder's inequality, see the proof of Lemma 1 in
\cite{li_schmidt-hieber_et_al_2024} for more details.

Having verified the GMC property \eqref{eq::gmc} with respect to the algorithmic
randomness introduced via the random weightings $D_k$, we may now integrate over
the distributions of $\whu_1$ and $\whv_1$, which proves geometric moment
contraction for all $q > 1$ such that the initial vectors admit a finite
$q$\textsuperscript{th} moment. Applying Corollary 4 of
\cite{li_schmidt-hieber_et_al_2024}, this proves existence and uniqueness of a
stationary distribution for the gradient descent iterates \eqref{eq::gd_rw}.
Further, this distribution is independent of the initialization.

To prove the final statement, let $\whv_\infty$ denote a random vector following
an independent copy of the stationary distribution. Initializing $\whweight_1
\sim \widehat{\mu}_1$ we now perform $k$ iterations \eqref{eq::gd_rw} on
$\whweight_1$ and $\whv_\infty$ with the same sequence $D_1, \ldots, D_k$ of
random weightings. Write $\whv_k$ for the iterates started from $\whv_\infty$.
Due to stationarity, $\whv_k$ induces the same measure $\widehat{\mu}_\infty$
for every $k$, so this yields a coupling of $\widehat{\mu}_k$ and
$\widehat{\mu}_\infty$. Together with the GMC property \eqref{eq::gmc} and the
definition of the transportation distance, this implies the convergence rate
\begin{align*}
  \calW^q_q\big(\widehat{\mu}_k, \widehat{\mu}_\infty\big) \leq \int
  \E_{D}\Big[\norm[\big]{\whweight_k - \whv_k}_2^q\Big]\ \rmd \widehat{\mu}_1
  \otimes \widehat{\mu}_\infty \leq C_3 \cdot \Big(1 - \alpha \cdot \big(2 -
  \alpha \tau^2 \cdot \norm{\bbX}\big) \cdot \sigminp(\whbbX)\Big)^k,
\end{align*} with constant $C_3 > 0$ implicit in the proof of Theorem 2,
\cite{wu_shao_2004}. Taking the $q$\textsuperscript{th} root and applying Lemma
\ref{lem::misc_a} now completes the proof.

\subsection{Proof of Theorem \ref{thm::conv_point}}

The triangle inequality implies $\calW_2\big(\widehat{\mu}_k,
\delta_{\whweight}\big) \leq \calW_2\big(\widehat{\mu}_k,
\widehat{\mu}_\infty\big) + \calW_2\big(\widehat{\mu}_\infty,
\delta_{\whweight}\big)$, so the result follows if both right-hand side
distances are bounded by $\eps / 2$. Applying Theorem \ref{thm::stat_dist}, the
first distance satisfies \begin{align*}
  \calW_2\big(\widehat{\mu}_k, \widehat{\mu}_\infty\big) \leq C_3 \cdot
  \exp\left(- \dfrac{\alpha \cdot \big(2 - \alpha \tau^2 \cdot \norm{\bbX}\big)
  \cdot \sigminp(\whbbX)}{2} \cdot k\right) < \dfrac{\eps}{2},
\end{align*} where the second inequality follows from plugging in the assumption
on $k$. We may estimate the second distance by bounding the expectation over a
particular coupling of $\widehat{\mu}_\infty$ and $\delta_{\whweight}$. Choosing
the product measure and letting $\whweight_\infty$ be as defined in
\eqref{eq::stat_vec}, this yields the bound \begin{align}
  \label{eq::w_bound}
  \calW_2^2\big(\widehat{\mu}_\infty, \delta_{\whweight}\big) \leq
  \E\Big[\norm[\big]{\whweight_\infty - \whweight}_2^2\Big]
\end{align} with $\E$ denoting the expectation over both the random weightings
$D_1, D_2, \ldots$ and the initializations. In particular, $\E_D$ then matches
the conditional expectation $\E\big[\ \cdot\ \mid \whweight_1\big]$. Recall from
Theorem 7.12 of \cite{villani_2003} that $\widehat{\mu}_k$ converging to
$\widehat{\mu}_\infty$ in $\calW_2$ implies $\lim_{k \to \infty}
\E\big[\norm{\whweight_k - \whweight}_2^2\big] = \E\big[\norm{\whweight_\infty -
\whweight}_2^2\big]$. By definition, $\norm{\bfv}_2^2 = \Tr\big(\bfv
\bfv\tran\big)$ and so continuity of the trace operator implies \begin{align*}
  \E\Big[\norm[\big]{\whweight_\infty - \whweight}_2^2\Big] = \lim_{k \to
  \infty} \E\Big[\norm[\big]{\whweight_k - \whweight}_2^2\Big] =
  \Tr\Bigg(\lim_{k \to \infty}  \E\Big[\big(\whweight_k - \whweight\big)
  \big(\whweight_k - \whweight\big)\tran\Big]\Bigg).
\end{align*} Using the limit computed in Theorem \ref{thm::var_conv_iso_b}, we
find the vanishing bound \begin{align*}
  &\lim_{k \to \infty} \norm*{\E\Big[\big(\whweight_k -
  \whweight\big) \big(\whweight_k - \whweight\big)\tran \bigmid
  \whweight_1\Big] - \big(\id - \Slin_\alpha\big)\inv_{\ker(X)}
  \big(\Sint_\alpha\big)}\\
  =\ &\lim_{k \to \infty} \norm*{\E\Bigg[\E\Big[\big(\whweight_k -
  \whweight\big) \big(\whweight_k - \whweight\big)\tran \bigmid
  \whweight_1\Big]\Bigg] - \big(\id - \Slin_\alpha\big)\inv_{\ker(X)}
  \big(\Sint_\alpha\big)}\\
  \leq\ &\lim_{k \to \infty} \E\Bigg[\norm[\bigg]{\E_D\Big[\big(\whweight_k -
  \whweight\big) \big(\whweight_k - \whweight\big)\tran\Big] - \big(\id -
  \Slin_\alpha\big)\inv_{\ker(X)} \big(\Sint_\alpha\big)}\Bigg]\\
  =\ &0,
\end{align*} which leads to the conclusion \begin{align*}
\E\Big[\norm[\big]{\whweight_\infty - \whweight}_2^2\Big] = \Tr\Bigg(\big(\id -
  \Slin_\alpha\big)\inv_{\ker(X)} \big(\Sint_\alpha\big)\Bigg).
\end{align*}

As shown in the proof of Theorem \ref{thm::var_conv_iso}, the inversion of the
linear operator $\id - \Slin_\alpha$ on the space of matrices with kernel
containing $\ker(X)$ results from the Neumann series $\sum_{\ell = 0}^\infty
(\Slin_\alpha)^\ell$. We also recall from Lemma \ref{lem::conv_exp_iso} that
\begin{align*}
  \Sint_{\alpha} = \alpha^2 \cdot X\tran \Bigg(\Sigma_D \odot \big(\bfY - X
  \whweight\big) \big(\bfY - X \whweight\big)\tran \Bigg) X,
\end{align*} which is a symmetric matrix with $\ker(X) \subset
\ker\big(\Sint_{\alpha}\big)$. As shown in \eqref{eq::slin_ker}, the kernel of
$\Slin_\alpha(A)$ contains $\ker(X)$ whenever $\ker(X) \subset \ker(A)$. Hence,
we may repeatedly apply Lemma \ref{lem::lin_op_bound} to find the estimate
\begin{equation}
  \label{eq::rem_est}
  \begin{split}
    \Tr\bigg(\big(\id - \Slin_\alpha\big)\inv_{\ker(X)}
    \big(\Sint_\alpha\big)\bigg) &\leq d \cdot \norm[\bigg]{\big(\id -
    \Slin_\alpha\big)\inv_{\ker(X)} \big(\Sint_\alpha\big)}\\
    &\leq d \cdot \sum_{\ell = 0}^\infty\
    \norm[\bigg]{\big(\Slin_\alpha\big)^\ell \big(\Sint_\alpha\big)}\\
    &\leq d \cdot \left(\sum_{\ell = 0}^\infty \big(1 - \alpha \cdot
    \sigminp(\whX)\big)^\ell\right) \cdot \norm[\big]{\Sint_\alpha} =
    \dfrac{d}{\alpha \cdot \sigminp(\whbbX)} \cdot \norm[\big]{\Sint_\alpha}.
  \end{split}
\end{equation} Lemma \ref{lem::pinv_b} and \ref{lem::misc_c}, as well as
sub-multiplicativity of the norm imply $\norm{\Sint_\alpha} \leq \alpha^2 \cdot
\norm{\Sigma_D} \cdot \norm{\bbX} \cdot \norm{\bfY - X \whweight}_2^2$.
Combining these computations into a bound for \eqref{eq::w_bound} and taking the
square root results in \begin{align*}
  \calW_2\big(\widehat{\mu}_\infty, \delta_{\whweight}\big) \leq \sqrt{\alpha
  \cdot \dfrac{d \cdot \norm{\Sigma_D} \cdot \norm{\bbX} \cdot \norm[\big]{\bfY
  - X \whweight}_2^2}{\sigminp(\whbbX)}}.
\end{align*} Due to the assumption on $\alpha$, the latter cannot exceed $\eps /
2$, which completes the proof.

\section{Proofs for Section \ref{sec::spec_sim}}

\subsection{Proof of Theorem \ref{thm::asym_srisk}}

Since $D_1, D_2, \ldots$ and $\bfeps$ are independent and the only sources of
randomness, we may write $\E_D\big[\ \cdot\ \big] = \E\big[\ \cdot\ \mid
\bfeps\big]$. The representation \eqref{eq::gd_var} of the gradient descent
recursion shows that $\whweight_k - \whweight$ is an affine function of $\bfY$,
which in turn also applies to $\starweight - \whweight_k$. Consequently,
$\norm{\starweight - \whweight_k}_2^2$ defines a polynomial of order $2$ in the
components of $\bfY$, with random coefficients that are independent of $\bfeps$.
The expectations of these coefficients must be uniformly bounded in $k$.
Otherwise, the convergent expression \eqref{eq::rec_unfold} in Theorem
\ref{thm::var_conv_iso_b} would diverge, thereby creating a contradiction. This
yields an integrable envelope, so dominated convergence implies \begin{align}
  \label{eq::lim_srisk}
  \lim_{k \to \infty} \E\Big[\norm[\big]{\starweight - \whweight_k}_2^2\Big] =
  \lim_{k \to \infty} \E\Bigg[\E\Big[\norm[\big]{\starweight - \whweight_k}_2^2
  \bigmid \bfeps\Big]\Bigg] = \E\Bigg[\lim_{k \to \infty}
  \E\Big[\norm[\big]{\starweight - \whweight_k}_2^2 \bigmid \bfeps\Big]\Bigg]
\end{align} Applying the convergence results in Lemma \ref{lem::conv_exp_iso}
and Theorem \ref{thm::var_conv_iso_b}, together with the continuous mapping
theorem (see Theorem 2.3 of \cite{van_der_vaart_1998}) the latter evaluates to
\begin{align}
  &\E\Bigg[\lim_{k \to \infty} \E\Big[\norm[\big]{\starweight - \whweight_k}_2^2
  \bigmid \bfeps\Big]\Bigg] \nonumber \\
  =\ &\E\Bigg[\lim_{k \to \infty} \norm[\Big]{\E\big[\starweight - \whweight_k
  \bigmid \bfeps\big]}_2^2 + \lim_{k \to \infty} \Tr\bigg(\Cov\big(\starweight -
  \whweight_k \mid \bfeps\big)\bigg)\Bigg] \nonumber \\
  =\ &\E\Big[\norm[\big]{\starweight - \whweight}_2^2\Big] + \E\Bigg[\lim_{k \to
  \infty} \Tr\Bigg(\E\Big[\big(\whweight_k - \whweight\big) \big(\whweight_k -
  \whweight\big)\tran \bigmid \bfeps\Big] - \E\big[\whweight_k - \whweight \mid
  \bfeps\big] \E\big[\whweight_k - \whweight \mid \bfeps\big]\tran\Bigg)\Bigg]
  \nonumber \\
  =\ &\E\Big[\norm[\big]{\starweight - \whweight}_2^2\Big] +
  \E\Bigg[\Tr\bigg(\big(\id - \Slin_{\alpha}\big)\inv_{\ker(X)}
  \big(\Sint_{\alpha}\big)\bigg)\Bigg]. \label{eq::srisk_decomp}
\end{align} Due to positive-definiteness, the trace in the previous display
cannot be negative, implying the bound $\E\big[\norm{\starweight -
\whweight}_2^2\big] \leq \lim_{k \to \infty} \E\big[\norm{\starweight -
\whweight_k}_2^2\big]$. As $\bfY = X \starweight + \bfeps$, this yields the
lower bound in Theorem \ref{thm::asym_srisk} via \begin{align}
  \E\Big[\norm[\big]{\starweight - \whweight}_2^2\Big] =
  \E\Big[\norm[\big]{\starweight - \whX\pinv \whY}_2^2\Big] &=
  \E\Big[\norm[\big]{\big(I - \whX\pinv \whX\big) \starweight - \whX\pinv
  M_2^{1 / 2} \bfeps}_2^2\Big] \nonumber \\
  &= \norm[\big]{\big(I - \whX\pinv \whX\big) \starweight}_2^2 +
  \Tr\Big(\big(\whX\pinv M_2^{1 / 2}\big) \Sigma_{\bfeps} \big(\whX\pinv M_2^{1
  / 2}\big)\tran\Big) \label{eq::srisk_wlls}
\end{align} where $I - \whX\pinv \whX = I - X\pinv X$ since the latter gives the
orthogonal projection onto $\ker(X) = \ker(\whX)$.

The corresponding upper bound follows via analysis of the trace term in
\eqref{eq::srisk_decomp}. Due to the specific form of $\Slin_\alpha$, as
computed in Lemma \ref{lem::var_op_iso}, $\Slin_\alpha(A)$ inherits symmetry
from $A$. Further, the Schur Product Theorem (see Theorem 5.2.1 of
\cite{horn_johnson_1991}) implies the same for positive semi-definiteness. Now,
using series expansion to invert $\id - \Slin_\alpha$, as well as applying Lemma
\ref{lem::lin_op_bound} together with Lemma \ref{lem::misc_d}, the trace term
satisfies \begin{align*}
  \Tr\bigg(\big(\id - \Slin_{\alpha}\big)\inv_{\ker(X)}
  \big(\Sint_{\alpha}\big)\bigg) &= \Tr\left(\sum_{\ell = 0}^\infty
  \big(\Slin_\alpha\big)^{\ell} \big(\Sint_\alpha\big)\right)\\
  &\leq \sum_{\ell = 0}^\infty \big(1 - \alpha \cdot
  \sigminp(\whbbX)\big)^{\ell} \cdot \Tr\big(\Sint_\alpha\big) =
  \dfrac{1}{\alpha \cdot \sigminp(\whbbX)} \cdot \Tr\big(\Sint_\alpha\big).
\end{align*} To further bound the trace in the previous display, note that $A
\mapsto \Sigma \odot A$ and $A \mapsto X\tran A X$ define linear operators that
satisfy the assumptions of Lemma \ref{lem::misc_d}. Consequently, since
$\norm{X} = 1$ \begin{align*}
  \dfrac{1}{\alpha \cdot \sigminp(\whbbX)} \cdot \Tr\big(\Sint_\alpha\big) &=
  \dfrac{\alpha}{\sigminp(\whbbX)} \cdot \Tr\Bigg(X\tran \bigg(\Sigma_D \odot
  \big(\bfY - X \whweight\big) \big(\bfY - X \whweight\big)\tran \bigg) X\Bigg)
  \\
  &\leq \dfrac{\alpha \cdot \norm{\Sigma_D}}{\sigminp(\whbbX)} \cdot
  \Tr\Big(\big(\bfY - X \whweight\big) \big(\bfY - X \whweight\big)\tran\Big) \\
  &\leq \dfrac{\norm{\Sigma_D}}{\sigminp(\whbbX)^2 + \norm{\Sigma_D}} \cdot
  \norm[\big]{\bfY - X \whweight}_2^2 
\end{align*} where the last inequality follows from the assumption on $\alpha$
in Theorem \ref{thm::var_conv_iso}. As $\norm{\Sigma_D} /
\big(\sigminp(\whbbX)^2 + \norm{\Sigma_D}\big) \leq 1$, it suffices to analyze
the expected squared norm of $\bfY - X \whweight$. By assumption, $\bfY = X
\starweight + \bfeps$ and so $X \big(I - \whX\pinv \whX\big) = 0$ implies
\begin{align*}
  \E\Big[\norm[\big]{\bfY - X \whweight}_2^2\Big] &= \E\Big[\norm[\big]{\big(X -
  X \whX\pinv \whX\big) \starweight + \big(I - X \whX\pinv M_2^{1 / 2}\big)
  \bfeps}_2^2\Big] \\
  &= \E\Big[\norm[\big]{\big(I - X \whX\pinv M_2^{1 / 2}\big)
  \bfeps}_2^2\Big]\\
  &= \Tr\Big(\big(I - X \whX\pinv M_2^{1 / 2}\big) \Sigma_{\bfeps} \big(I - X
  \whX\pinv M_2^{1 / 2}\big)\tran\Big).
\end{align*} Together with the computations \eqref{eq::srisk_decomp} and
\eqref{eq::srisk_wlls}, this yields the claimed upper bound for the right-hand
side of \eqref{eq::lim_srisk}.

\section{Auxiliary Results}

In this appendix we gather the main technical results used in the main text. Due
to the nature of the linear regression loss, these tools are predominantly based
on classical results in linear algebra.

\subsection{Singular Values and Pseudo-Inversion}
\label{sec::svalue_pinv}

Let $A$ be an $(n \times d)$-matrix, then any decomposition $A = U \Sigma
V\tran$ is a singular value decomposition if the following hold: \begin{enumerate}
  \item The matrix $U$ is orthogonal and of size $n \times n$.
  \item The matrix $V$ is orthogonal and of size $d \times d$.
  \item The matrix $\Sigma$ has non-negative diagonal entries, zeroes everywhere
    else, and is of size $n \times d$.
\end{enumerate} The diagonal entries of $\Sigma$ are unique and referred to as
the singular values of $A$. Note that $U = V$ for a symmetric square matrix.

We may use the singular value decomposition to invert $A$ on the largest
possible subspace. The pseudo-inverse $A\pinv$ of $A$ is given by $V \Sigma\pinv
U\tran$, where $\Sigma\pinv$ denotes the diagonal $(d \times m)$-matrix with
\begin{align*}
  \Sigma\pinv_{ii} = \begin{cases}
    \Sigma_{ii}\inv &\mbox{if } \Sigma_{ii} > 0, \\
    0 &\mbox{if } \Sigma_{ii} = 0.
  \end{cases}
\end{align*} By construction, $A\pinv A = V \Sigma\pinv \Sigma V\tran$ and $A
A\pinv = U \Sigma \Sigma\pinv U\tran$. Both $\Sigma\pinv \Sigma$ and $\Sigma
\Sigma\pinv$ are diagonal matrices, featuring binary diagonal entries. The
singular values of a non-singular matrix are positive, in which case
$\Sigma\pinv \Sigma = \Sigma \Sigma\pinv = I$ and so $A\pinv = A\inv$.

For more details, we refer to Chapter 17 of \cite{roman_2008}. We collect the
relevant properties of the pseudo-inverse in the following lemma.

\begin{lemma}
  \begin{enumerate}
    \item For any matrix $A$, the identity $A\tran A A\pinv = A\tran$ holds.
      \label{lem::pinv_a}
    \item For any vector $\bfv$ of compatible dimension, $A\pinv \bfv \perp
      \ker(A)$. \label{lem::pinv_b}
    \item The minimum norm minimizer of $\bfv \mapsto \norm{A \bfv - \bfw}_2$ is
      given by $A\pinv \bfw$. \label{lem::pinv_c}
    \item Given matrices $A$ and $B$ of compatible dimensions, $(A B)\pinv =
      B\pinv A\pinv$ whenever $A$ has linearly independent columns and $B$ has
      linearly independent rows. \label{lem::pinv_d}
  \end{enumerate}
\end{lemma}

\begin{proof}
  \begin{enumerate}
    \item Let $A = U \Sigma V\tran$ be a singular value decomposition, then
      $A\tran A A\pinv = V \Sigma\tran \Sigma \Sigma\pinv U\tran$. By
      construction, $\Sigma\tran \Sigma \Sigma\pinv = \Sigma\tran$ and the proof
      is complete.
    \item Note that $A\tran A = V \Sigma\tran \Sigma V\tran$, so $\bfw \in
      \ker(A) \subset \ker(A\tran A)$ implies $V\tran \bfw \in \ker(\Sigma\tran
      \Sigma)$. Since $\Sigma\tran \Sigma$ is diagonal, it follows that $(V\tran
      \bfw)_i$ can only be non-zero when $(\Sigma\tran \Sigma)_{ii} = 0$. By
      definition of $\Sigma\pinv$, this implies $\bfw\tran A\pinv \bfv =
      \bfw\tran V \Sigma\pinv U\tran \bfv = 0$.
    \item See Theorem 17.3 of \cite{roman_2008}.
    \item See \cite{greville_1966}.
  \end{enumerate}
\end{proof}

Lastly, we prove a result to estimate the convergence rate of fixed-point
iterations in terms of the non-zero singular values of a given matrix.

\begin{lemma}
  \label{lem::fixp_conv}
  Fix $A \in \R^{n \times d}$ and suppose $\norm{A\tran A} < 1$. If $\bfw \perp
  \ker(A)$, then also $(I - A\tran A) \bfw \perp \ker(A)$ and \begin{align*}
    \norm[\big]{\big(I - A\tran A\big) \bfw}_2 \leq \big(1 - \sigminp(A\tran
    A)\big) \cdot \norm{\bfw}_2
  \end{align*}
\end{lemma}

\begin{proof}
  Write $r = \rank(A)$ and let $A = U \Sigma V\tran$ be a singular value
  decomposition. Permuting the columns of $U$ and $V$, we may order the singular
  values such that $\Sigma_{ii} > 0$ if, and only if, $i \leq r$. By definition,
  the column vectors $\bfv_1, \ldots, \bfv_d$ of $V$ are orthogonal and in turn
  \begin{align*}
    I - A\tran A &= V \big(I - \Sigma\tran \Sigma\big) V\tran = \sum_{i = 1}^d
    \big(I - \Sigma\tran \Sigma\big)_{ii} \cdot \bfv_i \bfv_i\tran.
  \end{align*} The vector $\bfw$ may be expressed as a unique linear combination
  $\bfw = \sum_{i = 1}^d c_i \cdot \bfv_i$. Theorem 17.3 of \cite{roman_2008}
  shows that $\bfv_{r + 1}, \ldots, \bfv_d$ form an ortho-normal basis for
  $\ker(A)$, so $\bfw \perp \ker(A)$ implies $c_i = 0$ for all $i > r$. In
  turn, \begin{align*}
    \big(I - A\tran A\big) \bfw = \left(\sum_{i = 1}^{d} \big(I - \Sigma\tran
    \Sigma\big)_{ii} \bfv_i \bfv_i\tran\right) \left(\sum_{j = 1}^d c_j \cdot
    \bfv_j\right) = \sum_{i = 1}^r \big(I - \Sigma\tran \Sigma\big)_{ii} c_i
    \cdot \bfv_i,
  \end{align*} which proves that $(I - A\tran A) \bfw \perp \ker(A)$. Further,
  the assumption $\norm{A\tran A} < 1$ entails $0 < (\Sigma\tran \Sigma)_{ii} <
  1$ for every $i \leq r$, so the first $r$ diagonal entries of $I - \Sigma\tran
  \Sigma$ must all lie in $(0, 1)$. Accordingly, ortho-normality of the $\bfv_i$
  implies \begin{align*}
    \norm[\Big]{\big(I - A\tran A\big)\bfw}_2 &= \sqrt{\left(\sum_{i = 1}^r
    \big(I - \Sigma\tran \Sigma\big)_{ii} c_i \cdot \bfv_i\right)\tran
    \left(\sum_{j = 1}^r \big(I - \Sigma\tran \Sigma\big)_{jj} c_j \cdot
    \bfv_j\right)}\\
    &= \sqrt{\sum_{i = 1}^r \big(I - \Sigma\tran \Sigma\big)_{ii}^2 \cdot c_i^2}
    \leq \left(\max_{i = 1, \ldots, r} \big(1 - \Sigma\tran
    \Sigma\big)_{ii}\right) \cdot \norm{\bfw}_2.
  \end{align*} To complete the proof, it suffices to note that $(1 - \Sigma\tran
  \Sigma)_{ii}$ attains its maximum over $i = 1, \ldots, r$ at the smallest
  non-zero diagonal entry of $\Sigma\tran \Sigma$, which coincides with
  $\sigminp(A\tran A)$.
\end{proof}

\subsection{Miscellaneous Facts}

We collect various useful results in the following lemma.

\begin{lemma}
  \begin{enumerate}
    \item Let $c_i$, $i \geq 1$ be a sequence in $(0, 1)$, then
      \begin{align*}
        \prod_{i = 1}^k \big(1 - c_i\big) \leq \exp\left(- \sum_{i = 1}^k
        c_i\right).
      \end{align*} for every $k \geq 1$. In turn, $\limsup_{k \to \infty}
      \prod_{i = 1}^k (1 - c_i) = 0$ whenever $\sum_{i = 1}^\infty c_i =
      \infty$. \label{lem::misc_a}
    \item For any vectors $\bfu$ and $\bfv$, \begin{align*}
        \norm{\bfu \bfv\tran} = \norm{\bfu}_2 \norm{\bfv}_2.
      \end{align*} \label{lem::misc_b}
    \item Denote by $A \odot B$ the element-wise product of matrices, then
      \begin{align*}
        \norm{A \odot B} \leq \norm{A} \cdot \norm{B}.
      \end{align*} \label{lem::misc_c}
    \item For any symmetric positive semi-definite $d \times d$ matrix $A$ and
      linear operator $A$ that maps the space of such matrices to itself,
      \begin{align*}
        \Tr\big(S(A)\big) \leq \norm{S}_{\op} \cdot \Tr(A).
      \end{align*} \label{lem::misc_d}
  \end{enumerate}
\end{lemma}

\begin{proof}
  \begin{enumerate}
    \item Note that $0 < 1 - c_i < 1$ implies $\log(1 - c_i) \leq - c_i$.
      Consequently, \begin{align*}
        \prod_{i = 1}^k \big(1 - c_i\big) &= \exp \left(\sum_{i = 1}^k
        \log\big(1 - c_i\big)\right)\\
        &\leq \exp \left(- \sum_{i = 1}^k c_i\right).
      \end{align*}
    \item Given any unit vector $\bfw$ of the same dimension as $\bfv$, note
      that $\norm{\bfu \bfv\tran \bfw}_2 = \abs{\bfv\tran \bfw} \cdot
      \norm{\bfu}_2$. The inner product $\bfv\tran \bfw$ is maximized over the
      unit sphere by taking $\bfw = \bfv / \norm{\bfv}_2$, which proves the
      result.
    \item See Theorem 5.5.1 in \cite{horn_johnson_1991}.
    \item Let $A = U \Sigma U\tran$ be a singular value decomposition and recall
      that the trace operator satisfies $\Tr(A) = \sum_{i = 1}^d \Sigma_{ii}$.
      If $U_S \Sigma_S U_S\tran$ denotes a singular value decomposition of
      $S(A)$, then the definition of $\norm{S}_{\op}$ implies $\Sigma_{S, ii}
      \leq \norm{S}_{\op} \cdot \Sigma_{ii}$, provided the diagonal entries of
      both matrices are ordered in descending fashion. This completes the proof
      since \begin{align*}
        \Tr\big(S(A)\big) = \sum_{i = 1}^{d} \Sigma_{S, ii} \leq \norm{S}_{\op}
        \cdot \sum_{i = 1}^d \Sigma_{ii} = \norm{S}_{\op} \cdot \Tr(A).
      \end{align*}
  \end{enumerate}
\end{proof}

\end{document}